%% file: root.tex
\documentclass[journal]{IEEEtran}
\usepackage{amsmath,amsfonts}
\usepackage{amssymb}
\usepackage{algorithmic}
\usepackage{algorithm}
\usepackage{array}
\usepackage[caption=false,font=normalsize,labelfont=sf,textfont=sf]{subfig}
\usepackage{textcomp}
\usepackage{stfloats}
\usepackage{url}
\usepackage{verbatim}
\usepackage{graphicx}
\usepackage{cite}
\usepackage{cases}
\usepackage{multirow}
\usepackage{booktabs}
\usepackage{mathtools}
\usepackage{gensymb}

\graphicspath{{figures/}}

\newcommand{\bs}[1]{
        \boldsymbol{#1}
}

\begin{document}

\title{Using Intent Estimation and Decision Theory to Support Lifting Motions with a Quasi-Passive Hip Exoskeleton}

\author{Thomas Callens$^{1,2}$, Vincent Ducastel$^{3,4}$, Joris De Schutter$^{1,2}$, Erwin Aertbeli\"en$^{1,2}$.
        % <-this % stops a space
\thanks{This work was supported in part by the Research Foundation-Flanders (FWO) through a research grant (Exo4Work, SBO-E-S000118N) and by the Interreg North Sea Region Exskallerate Project }%
\thanks{$^{1}$Department of Mechanical Engineering, KU Leuven, Belgium {\tt \footnotesize thomas.callens@kuleuven.be}}%
\thanks{$^{2}$Flanders Make@KU Leuven}%
\thanks{$^{3}$Robotics and Multibody Mechanics Research Group, Vrije Universiteit Brussel, Belgium}%
\thanks{$^{4}$IMEC Brussels, Kapeldreef 75, 3001 Leuven, Belgium}}

\markboth{IEEE Transactions on Robotics}%
{Shell \MakeLowercase{\textit{et al.}}: A Sample Article Using IEEEtran.cls for IEEE Journals}

\IEEEpubid{0000--0000/00\$00.00~\copyright~2021 IEEE}


\maketitle

\begin{abstract}
This paper compares three controllers for quasi-passive exoskeletons.
The Utility Maximizing Controller (UMC) uses intent estimation to recognize user motions and decision theory to activate the support mechanism.
The intent estimation algorithm requires demonstrations for each motion to be recognized.
Depending on what motion is recognized, different control signals are sent to the exoskeleton.
The Extended UMC (E-UMC) adds a calibration step and a velocity module to trigger the UMC.
As a benchmark, and to compare the behavior of the controllers irrespective of the hardware, a Passive Exoskeleton Controller (PEC) is developed as well. 
The controllers were implemented on a hip exoskeleton and evaluated in a user study consisting of two phases.
First, demonstrations of three motions were recorded: squat, stoop left and stoop right.
Afterwards, the controllers were evaluated.
The E-UMC combines benefits from the UMC and the PEC, confirming the need for the two extensions.
The E-UMC discriminates between the three motions and does not generate false positives for previously unseen motions such as stair walking.
The proposed methods can also be applied to support other motions.
\end{abstract}

% As such, the E-UMC overcomes some drawbacks of passive exoskeletons.

\input{sections/introduction.tex}
\input{sections/related_work.tex}
\input{sections/hardware.tex}
\input{sections/methods.tex}
\input{sections/experiments.tex}

\input{sections/results.tex}
\input{sections/discussion.tex}
\appendix
\input{sections/appendix.tex}

\bibliographystyle{IEEEtran}

\bibliography{root.bib}
\vfill

\end{document}

%% file: sections/introduction.tex
\section{Introduction}
\label{sec: introduction}
% \textbf{Choose proper name for controllers: IRC, SPC, Extended IRC, or anything else and introduce it in abstract, introduction and overview figure}
% \noindent\rule{\columnwidth}{0.4pt}
% \textbf{Brainstorm on controller names}
% current: Intent Recognition Controller (IRC), Extended Intent Recognition Controller (EIRC)
% \begin{itemize}
%     \item Recognition and Decision Controller: R\&D controller, R-D controller etc.
%     \item Estimation-Based Decision controller: EBD controller, Motion Based Decision: MDB, MoBD etc.
%     \item Motion-Based support controller
%     \item Utility Maximizing Controller
% \end{itemize}
% Current: Simulating Passive Controller
% \begin{itemize}
%     \item Velocity Trigger Controller: VT 
%     \item Passive Exoskeleton controller
% \end{itemize}
% \noindent\rule{\columnwidth}{0.4pt}
According to the Eurofound study on working conditions and sustainable work, musculoskeletal disorders are one of the most common work-related health disorders in the European Union \cite{Eurofound2022}.
Moreover, about half of the workers reported to be working in tiring or painful positions.
Occupational exoskeletons aim to reduce loads for workers in industry during lifting tasks and are a promising technology to reduce prevalence of musculoskeletal disorders.
In particular, exoskeletons supporting the hips and lower back have gained interest from the research community as well as from industry, resulting in several prototypes being available on the market. \cite{Laevo,Backx,Crayx}.

A first class of exoskeletons are passive exoskeletons which provide support to a user through spring-based mechanisms.
Passive exoskeletons have been shown to reduce spinal muscle activities \cite{Alemi2019,Koopman2020,Madinei2020}. 
They are robust, provide reliable support and are lightweight.
Still, industry adoption of even simple passive exoskeletons remains limited (as noted in \cite{DeLooze2016}). 
Uncomfortable user experiences could be one possible explanation for this lack of adoption. 
Passive exoskeletons have two well-known drawbacks that lead to discomfort:
first, passive exoskeletons are not able to discriminate between different motion types. 
Hence, a user will always receive the same type of support, regardless of their intention.
Second, the support mechanism remains engaged even during motions that do not require support (e.g. walking).

As opposed to passive exoskeletons, active exoskeletons contain actuators to provide support to a user.
Active exoskeletons can mitigate the shortcomings of passive exoskeletons by providing support to a user based on the user's intention.
Similarly as for passive exoskeletons, active exoskeletons have been shown to reduce spinal muscle activations \cite{Li2021,Lanotte2021b,Heo2020}.
Although active exoskeletons look promising, they tend to be heavier and require more expensive actuators and high-capacity batteries \cite{Howard2020}.
Moreover, control strategies need to be implemented to control the actuators in these exoskeletons.
These control strategies should enable intuitive and comfortable user experiences.

Quasi-passive exoskeletons use a passive support mechanism that can be actively locked or unlocked using a clutch.
Such actuation systems are also called \textit{Clutched Elastic Actuators} \cite{Plooij2017}.
Because of this actuation mechanism, quasi-passive exoskeletons combine advantages of both passive and active exoskeletons.
Like passive exoskeletons, they are lightweight and robust due to the limited size and complexity of the actuators.
Like active exoskeletons, they provide more flexibility when supporting a user because the clutch mechanism can be locked or unlocked.

However, as is the case for active exoskeletons, quasi-passive exoskeletons need controllers to determine when support should be provided to a user.
Thus, an intention estimation algorithm should be developed to classify ongoing motions. 
The controller should subsequently use this classification when deciding to lock the support mechanism or not. 
% Use this command here to make sure column dus not overlap with ID at bottom of page.
\IEEEpubidadjcol

The main contribution of this work is the development of a controller for a quasi-passive exoskeleton that estimates a user's intent and subsequently decides whether support should be provided.
Demonstrations of specific motions are recorded and used to extract motion models. 
The motion models are stored in a database accessed by the controller when trying to recognize user motions.
% To recognize motions, the controller uses motion models stored in a database.
To decide whether support should be provided, the controller uses utility functions capturing the impact of the exoskeleton's operation on a user.
Since this controller maximizes a user's utility, it is denoted as the Utility Maximizing Controller (UMC).
To further stabilize the UMC, the controller was extended with a calibration step and uses an additional velocity-based trigger to switch the intention detection algorithm on and off, resulting in the Extended Utility Maximizing Controller (E-UMC).
The proposed methods were implemented on a quasi-passive hip exoskeleton to support three lifting motions: squat, stoop left and stoop right (see bottom of figure \ref{fig: side and back view}). 
The resulting controllers were evaluated in a user study and compared with a benchmark controller emulating a passive exoskeleton (Passive Exoskeleton Controller, PEC).

The remainder of this paper is structured as follows. 
Section \ref{sec: related work} highlights related work.
Section \ref{sec: exoskeleton hardware} introduces the exoskeleton hardware prototype while section \ref{sec: methods} presents the control methods.
Sections \ref{sec: user study} and \ref{sec: results} discuss the user study and present results obtained.
Finally, the paper ends with a discussion and a conclusion section.

%% file: sections/related_work.tex
\section{Related work}
\label{sec: related work}
The literature on human-robot interaction contains several approaches for intent estimation.
Often, a set of motion models is created with which an ongoing user motion can be compared.
The most appropriate motion model is then selected and used to generate a corresponding robot motion \cite{Maeda2017a,Gaspar2018}.
A similar strategy can be followed when developing control strategies for exoskeletons.

Several approaches have been proposed to learn such motion models.
Dynamic Movement Primitives (DMPs) are among the most well-known methods to generate motion models \cite{DMP2013}. 
Apart from usage in intent estimation algorithms, DMPs have been used to generate support trajectories for exoskeletons as well \cite{Lanotte2021b,Huang2016}.
A disadvantage of DMPs is the lack of a probabilistic framework.
When trying to recognize ongoing user motions, a DMP-based algorithm is not able to communicate any uncertainties.

Probabilistic approaches to learning motion models have been proposed as well, such as Probabilistic Movement Primitives (ProMPs) \cite{Paraschos2018},
Probabilistic Principal Component Analysis (PPCA) \cite{DeSchutter2014} or Gaussian Mixture Models (GMMs) \cite{Luo2018}. 
These probabilistic approaches have been used in a variety of applications ranging from coordination tasks between human and robot to estimation of gait trajectories for lower-limb exoskeletons e.g. \cite{Perico2019,Maeda2017,Tanghea}.
Earlier work compared ProMPs and PPCA models when used to recognize and predict ongoing user motions \cite{Callens2020}. 

In human-exoskeleton interaction, in particular in interactions with quasi-passive or active exoskeletons, there is a need to estimate a user's intent as well.
% Quasi-passive hip exoskeletons have recently gained interest in the research community due to their limited hardware complexity as compared to active exoskeletons.
Quasi-passive exoskeletons were developed to provide support during squat motions \cite{Jamsek2020b,Wang2021,Hassan2019,Perera2020b}.
Other devices focused on supporting walking motions \cite{Shafer2022} or elbow lift motions \cite{Winter2021}.

Jam\u{s}ek et al. developed a quasi-passive hip exoskeleton \cite{Jamsek2020b} to support squatting and lifting motions.
Although the intention estimation algorithm, which uses Gaussian Mixture Models, was able to recognize and support squat and lift motions, it was not evaluated for asymmetric lift motions.
Moreover, the exoskeleton provided the same type of support for each recognized motion.
The HipExo developed by Perera et al. uses ad-hoc rules to detect forward bending motions and is not able to discriminate between asymmetric lifting and symmetric lifting \cite{Perera2020b}.

Apart from quasi-passive exoskeletons, active exoskeletons also need control strategies to support a user's motion.
Often, lifting motions are detected by rule-based intention estimation algorithms, such as in \cite{Ko2018,Li2021}.
Lanotte et al. used DMPs to generate a prediction of the human lifting motion and subsequently used this prediction to provide support to a user \cite{Lanotte2021b}.
While these approaches work well, it becomes difficult to discriminate between symmetric and asymmetric lifting motions. 
Active exoskeletons aimed at supporting walking motions usually start from the assumption that a user is walking, as in \cite{Giovacchini2015b,Nalam2022,Pour2022}.
Hence, no intention estimation algorithm is necessary.

This work proposes controllers which make use of PPCA motion models to recognize ongoing motions. 
With these motion models, the exoskeleton is able to estimate a user's intention. 
Next, the controller leverages the uncertainty obtained during the intent estimation step when deciding to provide support.
In this decision step, the controller takes into account the benefits and drawbacks of making a correct or wrong decision.
Depending on the recognized motion and uncertainty with which a motion is recognized, the controller sends different control signals to the actuation mechanisms.

% \noindent\rule{\columnwidth}{0.4pt}
% \begin{itemize}
%     \item Passive exoskeletons: not really much to say about this? Perhaps leave it out
%     \item Active exoskeletons
%     \item Quasi-passive exoskeletons
%     \item Focus a bit on control strategies of active and quasi passive exoskeletons in above references
%     \item PPCA and its applications
%     \item Any other relevant works? Maybe intent recognition in robotics applications
% \end{itemize}
% \noindent\rule{\columnwidth}{0.4pt}
%%% Local Variables:
%%% mode: latex
%%% TeX-master: "../root"
%%% End:

%% file: sections/hardware.tex
\section{Hardware}
\label{sec: exoskeleton hardware}
\begin{figure}
    \centering
    \includegraphics[width=0.7\columnwidth]{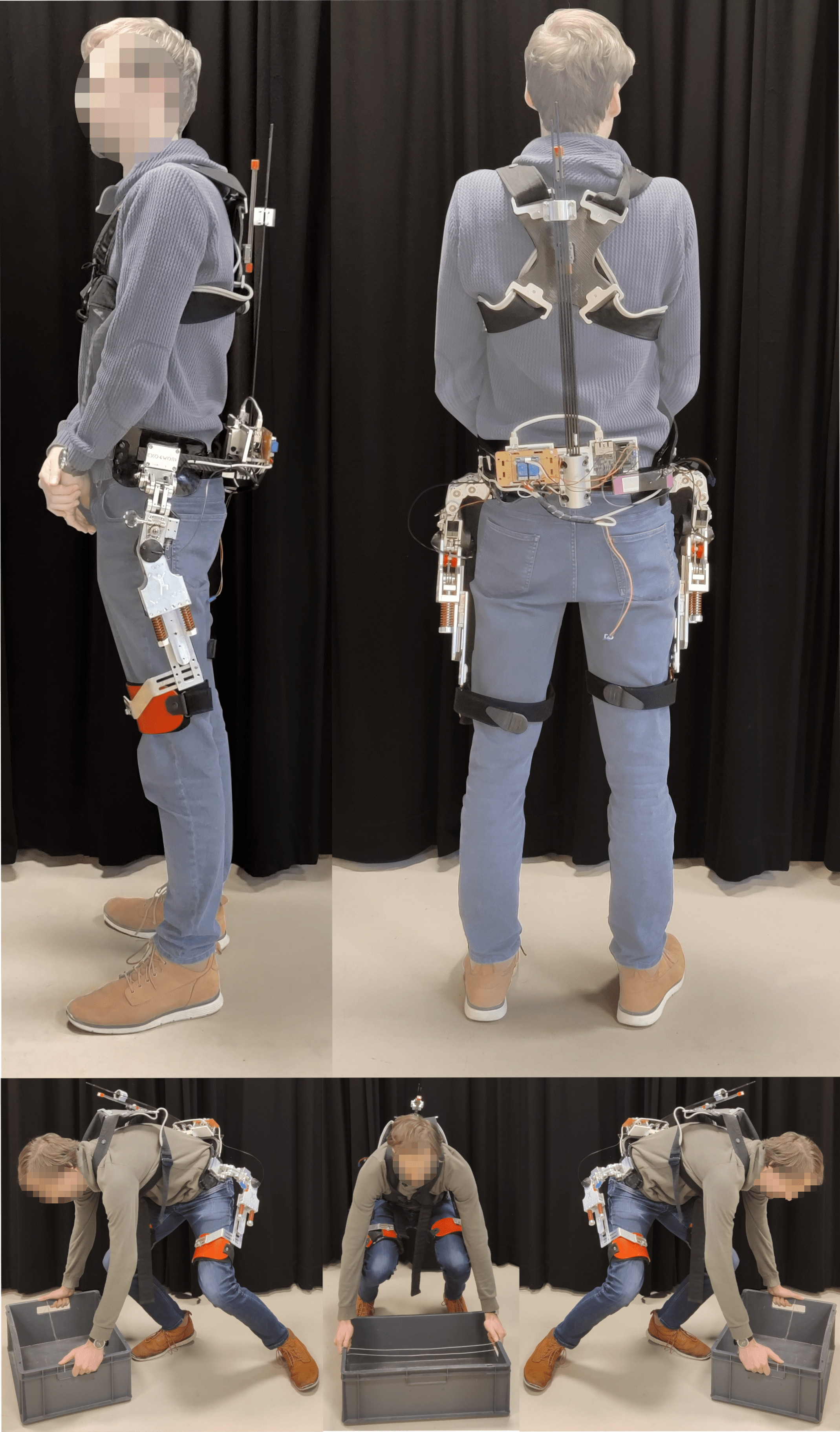}
    \caption{\textit{Top:} A side and back view of the quasi-passive exoskeleton.
    \textit{Bottom:} The three lifting motions to be recognized by the exoskeleton: Asymmetric stoop right, squat and asymmetric stoop left.}
    \label{fig: side and back view}
\end{figure}

Figure \ref{fig: side and back view} shows the current quasi-passive exoskeleton prototype, which is a modified version of the one proposed in \cite{Naf2018a}. 
The bottom part of the figure shows the three lifting motions on which the proposed methods are applied.

\begin{figure}
\centering
\includegraphics[width=0.8\columnwidth]{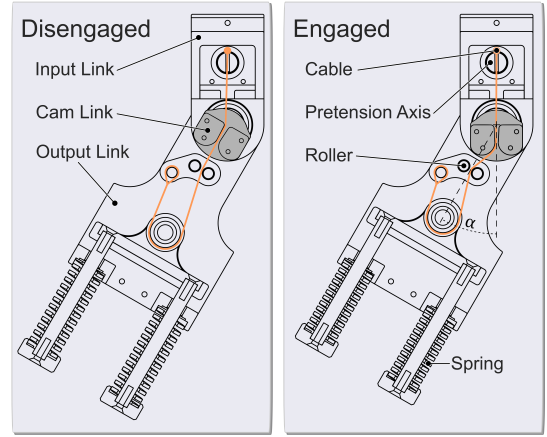}
\caption{Cross-section view of the left hip module presenting the operation principle of the actuator. 
When disengaged, the cam link (gray) follows the motion of the output link without compressing the springs. 
When engaged, the cam link is grounded to the input link. 
Any subsequent hip flexion will increase $\alpha$ which results in torque being generated by the springs. }
\label{fig: principle}
\end{figure}

% The new hip module uses a locking mechanism to allow engaging the springs only when required, so that no support is provided if it is not necessary.
% The operation principle of the locking mechanism is shown in Figure \ref{fig: principle}.
% The cam system from \cite{Naf2018a} has been uncoupled from the output link into an independent cam link which can be connected to the input whenever needed.
% If support is required, the cam link is locked with respect to the input link.
% Any subsequent hip flexion rotates the output link making the rollers pull the cable onto the cam. 
% This compresses the springs and creates a lever arm, resulting in a torque which assists the user to come back upright.
% However, if no support is required (e.g. when the user is walking), the cam link is disengaged and follows the motion of the output without pulling on the rope.
The new hip module uses a locking mechanism to allow engaging the springs only when required, so that no support is provided if it is not necessary.
The operation principle of the locking mechanism is shown in Figure \ref{fig: principle}.
The cam system from \cite{Naf2018a} was uncoupled from the output link into an independent cam link which can be connected to the input whenever needed.
If support is required, the cam link is locked with respect to the input link.
Any subsequent hip flexion will increase the deflection angle $\alpha$ between cam link and output link (see right part of figure \ref{fig: principle} where $\alpha>0$).
This in turn makes the rollers pull the cable onto the cam which compresses the springs and creates a lever arm, resulting in a torque that attempts to realign the output link with the cam link (such that $\alpha=0$).
When no support is required (e.g. when the user is walking), the cam link is disengaged and follows the motion of the output without pulling on the rope (thus, $\alpha=0$).

If $\theta$ denotes the hip flexion angle and if $\theta_{\text{lock}}$ denotes the angle at which the cam link was locked (in figure \ref{fig: principle}, $\theta_{\text{lock}}=0\degree$), then the deflection angle $\alpha$ is calculated as:
\begin{equation}
    \label{eq: deflection angle}
    \alpha = \theta - \theta_{\text{lock}}
\end{equation}
The rope used is a $\emptyset$2mm \textit{Dyneema} (breakload: 4100N) which offers high load capability at limited elongation.
To reduce the width of the hip module, the bulky spring of \cite{Naf2018a} was replaced by two smaller springs in parallel (SODEMANN ST51590, K = 33.6N/mm).

\begin{figure}
\centering
\includegraphics[width=0.8\columnwidth]{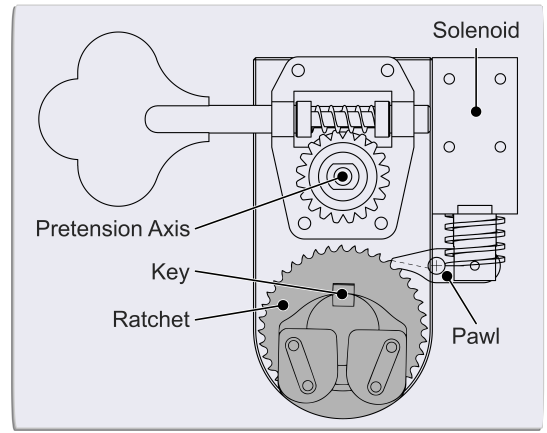}
\caption{Side view of the left side locking mechanism used to engage the springs.
The ratchet is connected to the cam link (in dark gray) via a key.
The solenoid is attached to the input link.
When activated, the solenoid pulls on the pawl which locks the ratchet. 
The mechanism is designed to be self-locking such that current only needs to be applied during the locking phase.}
\label{fig: locking}
\end{figure}

The locking system consists of a ratchet and pawl mechanism, as illustrated in Figure \ref{fig: locking}.
The pawl, connected to the input link, is actuated by a solenoid (RS PRO 177-0139, 12V), whereas the ratchet is attached to the cam link.
When current is applied, the solenoid pulls the pawl and locks motion of the ratchet.
Any subsequent clockwise rotation of the cam link increases $\alpha$ and compresses the springs which keeps the system locked so that no current needs to be applied to keep the pawl in the locking position.
% Any subsequent hip flexion puts the springs under tensions which keeps the system locked until the end of the task i.e. when the user is back in the position they were in at activation of the solenoid.
% Additionally, this mechanism is self-locking so that no current needs to be applied to keep the pawl in the locking position.
This also prevents unlocking halfway through a motion which could lead to a potentially dangerous release of the potential energy in the springs.
The ratchet possesses 40 teeth, therefore offering a position accuracy of 9\textdegree.
A spring is mounted on the solenoid that pushes the pawl away from the ratchet if $\alpha=0$. 
This ensures that the support mechanism will unlock automatically at the end of each motion.

The ratchet of the mechanism only remains locked if $\alpha>0$.
In figures \ref{fig: principle} and \ref{fig: locking}, showing the left side mechanism, this occurs with clockwise rotations of the output link with respect to the cam link after locking the solenoid.
During counterclockwise rotations, locking the solenoid will not block the ratchet and will not result in torques generated by the spring mechanism.
On the exoskeleton, clockwise rotation of the left side mechanism corresponds to hip flexion motions.
Thus, the exoskeleton can only build up support if the solenoid is locked during the hip flexion phase of a motion.
The same conclusion is true for the right side support mechanism.


\begin{figure}
\centering
\includegraphics[width=\linewidth]{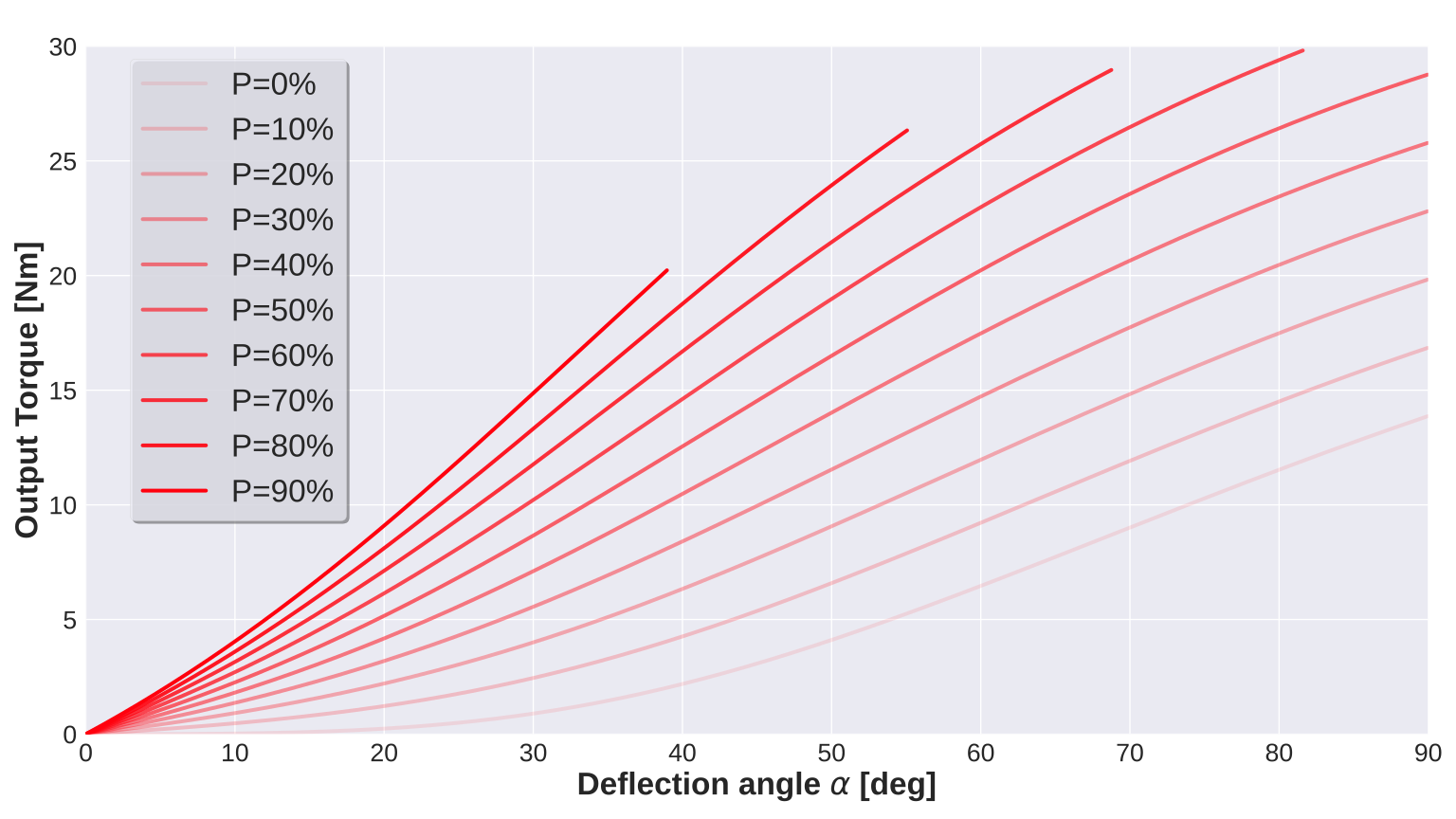}
\caption{Assistive torque provided by one hip module as function of the deflection $\alpha$ (angle between the output link and the cam link) for several levels of spring precompression $P$.
The spring precompression is reported as percentage of the maximal compression of the spring which is 22mm.}
\label{fig: torqueDeflection} 
\end{figure}

The precompression of the springs can be changed by increasing the pretension of the rope.
Indeed, the rope is wrapped around a pretension axis which is connected to a commercially available \textit{Schaller} bass tuner.
The assistive torque provided by the hip module can be estimated using the MACCEPA 2.0 model of \cite{Vanderborght2011c}, given some dimensions of the system which are reported in Table \ref{tab: MACCEPA}.
The torque curves generated as function of the deflection angle $\alpha$ (see Figure \ref{fig: principle}), are plotted in Figure \ref{fig: torqueDeflection} for several levels of spring precompression (reported as a percentage of the maximal spring compression which is 22mm).

\begin{table}
\caption{Main Parameters of the Actuator\label{tab: MACCEPA}}
\centering
\begin{tabular}{l c}
\toprule
\textbf{Parameter} & \textbf{Value [mm]} \\
% \hline
% \hline
\midrule
B : Length of the Lever Arm & 10\\
%\hline
C : Distance between Rollers and Rotation Point & 30\\
%\hline
R : Radius of the Profile Disk & 7.5\\
\toprule
% \hline
\end{tabular}
\end{table}

To estimate the state of the exoskeleton, the angular positions of the cam link and output link are measured by two magnetic absolute encoders (AMS AS5048A, SPI type, 6pins, 5V, 14bits, 0.0219deg/c).

%%% Local Variables:
%%% mode: latex
%%% TeX-master: "../root"
%%% End:

%% file: sections/methods.tex
\section{Controller methods}
\label{sec: methods}
This section introduces the methods and controllers used during the user study.
% Subsection \ref{sec: requirements} presents the controller requirements.
Subsections \ref{sec: controller overview} to \ref{sec: velocity trigger} present the three proposed controllers while subsection \ref{sec: practical implementation} discusses the practical implementation of these controllers.

\subsection{Utility Maximizing Controller}
\label{sec: controller overview}
\begin{figure*}[t]
    \centering
    \includegraphics[width=\textwidth]{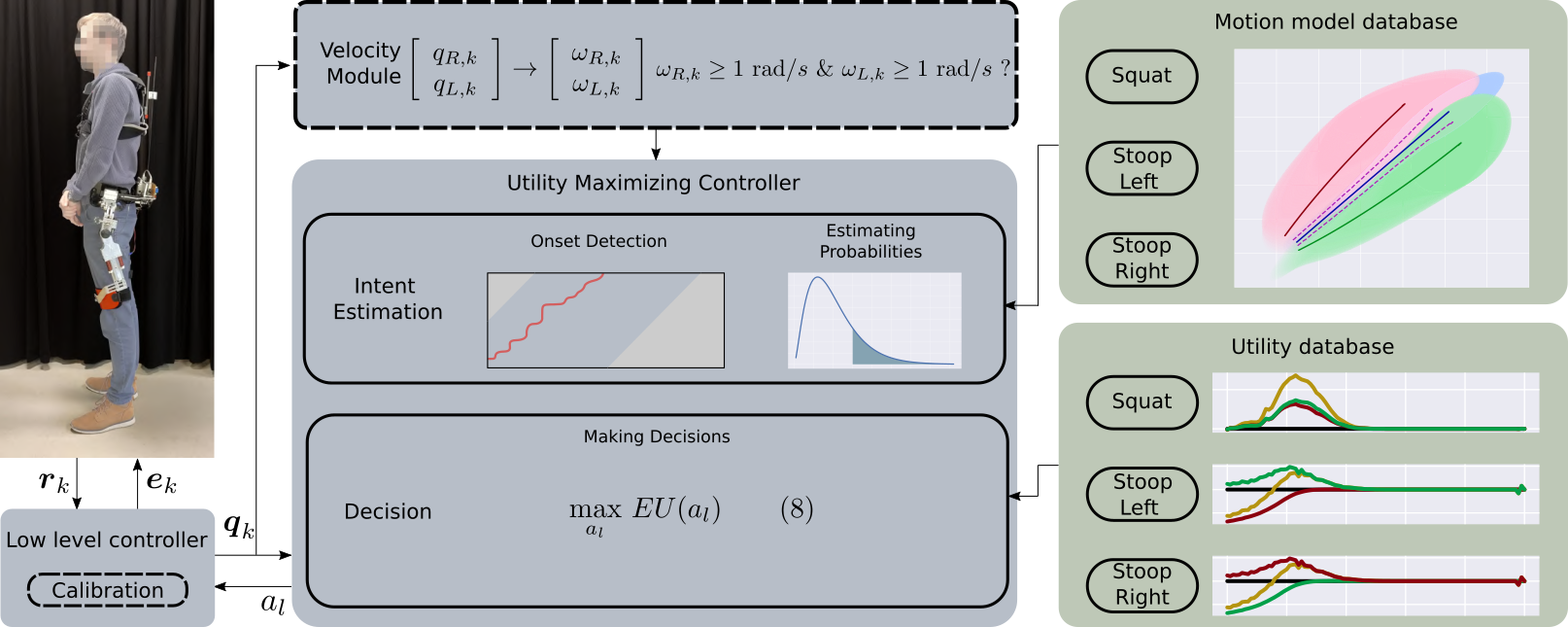}
    \caption{Overview of the (Extended) Utility Maximizing Controller. The subscript $k$ refers to the discrete time index.
    Raw encoder data is sent from the exoskeleton to a low-level controller where hip flexion angles $\bs{q}_k$ are calculated.
    In the intent estimation step, the controller will first perform onset detection and subsequently estimate model probabilities.
    In the decision step, the controller decides to take the action that maximizes utility to a user.
    The green blocks represent model data that is loaded at startup and used at runtime by the controller.
    The blocks in dashed lines (the calibration and velocity modules) are the extensions added to the Utility Maximizing Controller to arrive at the Extended Utility Maximizing Controller.}
    \label{fig: controller_overview}
\end{figure*}

Figure \ref{fig: controller_overview} shows an overview of the (Extended) Utility Maximizing Controller of the hip exoskeleton.
At each time instance $k$, the exoskeleton outputs an array of raw encoder data $\boldsymbol{r}_k$ for the four encoders mounted on the exoskeleton.
The low-level controller converts this encoder data to an array of two joint angles $\boldsymbol{q}_k$ that is sent to a buffer accessible by the UMC.
The UMC compares the data stream $\boldsymbol{q}_k$ with the motion models in the database.
The result of this intent estimation step is a probability estimate for each motion model.
In the decision step, the controller evaluates whether the probability assigned to a certain motion model is sufficient to lock the support mechanism on the left and/or the right side.
Decision theory is used to take the impact of potentially good and wrong decisions into account.
The controller sends a command signal $a_{l}$ back to the low-level controller, which converts it to an array $\boldsymbol{e}_k$ of two signals, one for each side of the exoskeleton.
The low-level controller holds the signal $\bs{e}_k$ for 0.3s to ensure that the ratchet and pawl of the support mechanism are properly locked.

The control signal $a_{l}$ can take on four possible values:
\begin{itemize}
    \item[$a_1$] do not lock
    \item[$a_2$] lock both sides
    \item[$a_3$] lock right side
    \item[$a_4$] lock left side
\end{itemize}
% \begin{itemize}
    %     \item[$k$] Time index at runtime
    %     \item[$n$] Progress index with $n=0\ldots N$ and $N=101$
    %     \item[$i$] Joint angle index in window of observations with $i=1\ldots30$
    %     \item[$j$] Motion model index with $j=1\ldots 3$
    %     \item[$l$] Control signal index with $a=1\ldots 4$
    % \end{itemize}
\begin{table}
    \centering
    \caption{Overview of subscripts}
    \label{tab: subscripts}
    \begin{tabular}{ll}
        \toprule
        Subscript & Meaning \\
        \midrule
        $k$ & Time index \\
        $n$ & Progress index with $n=0\ldots N$ and $N=101$ \\
        $i$ & Index in window of observations with $i=1 \ldots 30$ \\
        $j$ & Motion model index with $j=1 \ldots 3$ \\
        $l$ & Control signal index with $l=1\ldots 4$ \\
        \bottomrule
    \end{tabular}
\end{table}
The meaning of the subscripts used throughout this section is summarized in table \ref{tab: subscripts}.
The following subsections discuss the different blocks in the controller from figure \ref{fig: controller_overview}.

% \subsection{Low level controller}
% \label{sec: low level controller}
% As can be seen from figure \ref{fig: controller_overview}, the Utility Maximizing Controller interacts with a low level controller which in turn interacts with the exoskeleton.
% The low level controller converts encoder signals to joint angles.
% While the UM controller is evaluating the intent estimation and decision algorithms, the low level controller stores joint angle values in a buffer.
% This ensures that on the next iteration of the UM controller, the most recent joint angle data is available.

% Second, the Utility Maximizing Controller sends out a pulse $a_k$ whenever the decision algorithm has been evaluated.
% However, to make sure the support mechanism is properly locked, the low level controller holds this pulse for 0.3s (signal $\bs{e}_k$ in figure \ref{fig: controller_overview}).
% If the lock signal is sent during a lifting motion, then holding the lock signal ensures that a user has sufficiently flexed the hips to activate the self-locking mechanism in the exoskeleton.

\subsubsection{Learning motion models}
\label{sec: motion models}
The motion models aim to capture the variability displayed by exoskeleton users while executing specific motions.
If the motion models can capture this variability, they will be better suited to recognize motions from various exoskeleton users.
In this work, the PPCA methodology is used to learn motion models.
Details on the use and learning of PPCA motion models can be found in \cite{Callens2020,DeSchutter2014} and a more in-depth discussion can be found in \cite{Bishhop}.

To create motion models for the hip exoskeleton used here, users first have to perform squat, stoop left and stoop right motions while wearing the exoskeleton.
Throughout each sequence of demonstrations the encoders on the exoskeleton are continuously registering the hip flexion angles.
This sequence of demonstrations is subsequently segmented into individual lifting motions and the segmented demonstrations are resampled such that each demonstration contains an equal number of samples $N$.
Finally, the time value of each demonstration is replaced with a progress value $s_n$ which is defined to increase linearly with time such that $s_0=0$ at the start of the motion and $s_N=1$ at the end of the motion (with $N$=101).

These segmented, time-normalized and resampled demonstrations are used to learn three PPCA models following the procedure from \cite{DeSchutter2014}.
% This results in a PPCA model that describes the left and right hip flexion at a given progress value during a lifting motion with a normal distribution \textbf{Discuss index 'j' representing the model here}:
This results in a PPCA model$_j$ (with $j=1,2,3$) for each lifting motion describing the left and right hip flexion at a given progress value with a normal distribution:

\begin{equation}
    \label{eq: joint distribution}
    \bs{f}_j(s_n) \sim N(\bs{\mu}_{j}(s_n), \Sigma_{j}(s_n))
\end{equation}
with $\bs{\mu}_j(s_n)$ and $\bs{\Sigma}_j(s_n)$ corresponding to the mean and covariance of the demonstrations for model$_j$ at progress $s_n$.
Note that, in contrast to \cite{DeSchutter2014}, the latent variables of the PPCA model are here kept constant and assumed to be zero but the mean $\bs{\mu}_j$ and covariance $\Sigma_j$ are used further on in this work.

The three motion models are stored in the motion model database.
Since the solenoid in the support mechanism has to be locked during the hip flexion part of each motion, only this part of the motion models is stored for use in the UMC. 
For the three lifting motions, this corresponds to the first half of each motion model.

\subsubsection{Onset Detection}
\label{sec: onset detection}
In the onset detection step, the exoskeleton data stream is compared with the motion models in the database.
In the UMC, the aim is to detect when a lifting motion has started.
Thus, the algorithm is performing online segmentation.

To detect motion onsets, the well-known Dynamic Time Warping algorithm (DTW) is used \cite{Sakoe1978}.
Here, DTW is used to find an alignment between an ongoing time series and a part of a model time series.
The ongoing time series consists of a window of left and right hip joint angles over a horizon of 0.3s and sampled at 100Hz.
From here on, this window will be referred to as the ``observations'' and will be denoted as $\text{\textbf{obs}}= \left[\bs{q}_{1} \ldots \bs{q}_{i},\ldots,\bs{q}_{30}\right]$.
The model time series is a time-normalized motion model from the database.
Thus, to compare the observations with all the motion models the algorithm below is repeated for every motion model.

\begin{figure}
    \centering
    \includegraphics[width=0.7\columnwidth]{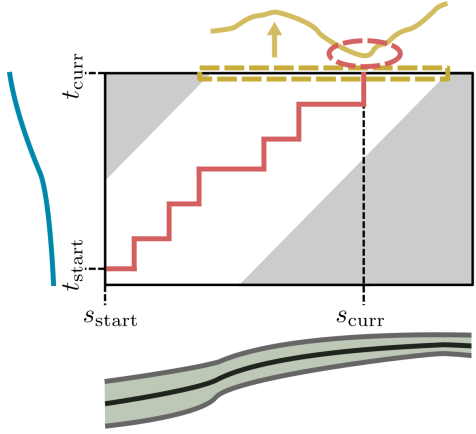}
    \caption{This figure shows a 1D representation of the DTW algorithm.
    The rectangle represents the DTW matrix.
    \textit{Green}: the motion model used.
    \textit{Blue}: the observations.
    \textit{Yellow}: values in the top row of the DTW matrix.
    \textit{Red}: the warping function.
    }
    \label{fig: dtw process}
\end{figure}

Similarly as in \cite{Sakoe1978}, an adjustment window is used when creating the DTW matrix to limit the computational load of the algorithm.
Figure \ref{fig: dtw process} visualizes the DTW matrix as a rectangle with the adjustment window shown in white.

A few modifications were made to the originally proposed DTW algorithm.
First, the boundary conditions were relaxed to allow the warping function to start anywhere on the bottom or left edge of the DTW matrix inside the adjustment window and end anywhere on the top edge of the DTW matrix inside the adjustment window (i.e. start and end on the edges inside the white area of figure \ref{fig: dtw process}).
Second, most DTW applications use a squared Euclidean distance as distance metric to build the DTW matrix.
% The current implementation leverages the probabilistic nature of the motion models by using squared Mahalanobis distances to the PPCA motion model at every element of the DTW matrix.
Instead, the current implementation leverages the probabilistic nature of the motion models by using squared Mahalanobis distances. % at every element of the DTW matrix.
The squared Mahalanobis distance $M$ between observation sample $\bs{q}_i$ and model$_j$ at the discrete progress value $s_n$ then becomes:
\begin{equation}
    \resizebox{0.91\hsize}{!}{$M\left[\bs{q}_i, \bs{f}_j(s_n)\right] = (\bs{q}_i-\bs{\mu}_{j}(s_n))^T\bs{\Sigma}_{j}(s_n)^{-1}(\bs{q}_i-\bs{\mu}_{j}(s_n))$}
\end{equation}

After the DTW matrix has been built, the top row of the matrix contains cumulative Mahalanobis distances.
$D_j$ corresponds to the minimum value of this top row of the DTW Matrix and will be referred to as the DTW distance of the observations to model$_j$.
The optimal alignment between the observations and model$_j$ is then extracted by retracing the warping function starting from the minimum in the top row.

Figure \ref{fig: dtw process} summarizes this procedure.
The adjustment window of the DTW matrix is filled in according to \cite{Sakoe1978} using squared Mahalanobis distances between the observations (blue) and model$_j$ (green).
Values in the top row of the DTW matrix are given by the yellow curve and the DTW distance $D_j$ is encircled in red.
The warping function corresponding to this distance $D_j$ is shown in red and is retraced starting from the element of the DTW matrix containing the value $D_j$.
This discrete warping function has a length $L_j$.
% The warping function does not necessarily include all available observations $\bs{q_i}$ or model samples $\bs{f}(s_j)$
The time values of the first and most recent observations included in the warping function (indicated with $t_{\text{start}}$ and $t_{\text{curr}}$) as well as the progress values of the first and last model samples included in the warping function (indicated with $s_{\text{start}}$ and $s_{\text{curr}}$) are visualized in figure \ref{fig: dtw process} as well.
Note that $s_{\text{curr}}$ is an estimate of the progress value of the most recent observation with respect to the motion model i.e. it is an estimate of the current progress of a user with respect to the motion model.

Similarly as in \cite{Callens2020}, a match between the observations and the motion model must satisfy three rough criteria.
First, $s_{\text{curr}}-s_{\text{start}} > s_{\text{req}}$ requires that at least a fraction $s_{\text{req}}$ of the motion model is used in the onset detection procedure.
Second, $t_{\text{end}}-t_{\text{start}} > t_{\text{req}}$ requires that the observation samples used in the onset detection procedure span at least $t_{\text{req}}$ seconds.
Third, $D_j < D_{\text{max}}$ immediately discards matches with extremely high DTW distances.
$D_{\text{max}}$ is set as the cumulative Mahalanobis distance arising from a match with warping function length $=$ 30 in which all observations lie $2\bs{\sigma}_j$ from the mean of the motion model $\bs{\mu}_j$.

As the onset detection algorithm is repeated for each model$_j$, different DTW distances $D_j$, progress estimates $s_{\text{curr},j}$ and warping functions with lengths $L_j$ are generated for each model.

\subsubsection{Estimating model probabilities}
\label{sec: motion probabilities}
This section discusses how to calculate the posterior probability $P(\text{model}_j|\text{\textbf{obs}})$ given the output of the onset detection algorithm.
If a motion does not correspond to a model from the database, it must be classified as ``other'' motions.
We therefore assume that only four options are possible: squat, stoop left, stoop right and other.

The DTW distance $D_j$ returned by the onset detection algorithm can be interpreted as a measure of the agreement between the observations and model$_j$.
Therefore, $D_j$ is used to calculate the probability of the observations given model$_j$ i.e. $P(\text{\textbf{obs}}|\text{model}_j)$.
A low DTW distance $D_j$ should result in a high probability and a high distance $D_j$ should result in a low probability.
An extremely low value $D_j$ indicates an almost perfect agreement between observations and model$_j$ and should result in a very high probability $P(\text{\textbf{obs}}|\text{model}_j)$.
Since $D_j$ is calculated as the sum of $L_j$ 2-dimensional Mahalanobis distances, it follows a $\chi^2$ distribution with $2L_j$ degrees of freedom. 
The probability $P(\text{\textbf{obs}}|\text{model}_j)$ is thus calculated using the $\chi^2$ distribution:
% The probability of observing a Mahalanobis distance $D$ or higher for a given $\text{model}_j$ is written as $P(D^{+}|\text{model}_j)$.
% Since $D$ is calculated as the sum of $L$ 2-dimensional Mahalanobis distances, it follows a $\chi^2$ distribution with $2L$ degrees of freedom.
% Therefore, $P(D^{+}|\text{model}_j)$ is given by:

\begin{figure}
    \centering
    \includegraphics[width=\columnwidth]{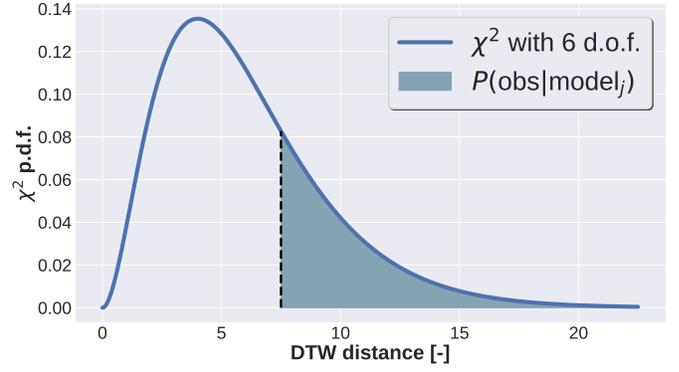}
    \caption{Figure showing the probability density function (p.d.f) of a 6 d.o.f. $\chi^2$ distribution.
    If the onset detection algorithm returns a DTW distance $D_j=7.5$ then $P(\text{\textbf{obs}}|\text{model}_j)$ is given by the shaded area.}
    \label{fig: chi squared}
\end{figure}

% Since the DTW distance $D$ is calculated as the sum of $L$ 2-dimensional Mahalanobis distances it follows a chi-square distribution with $2L$ degrees of freedom i.e. $D\sim\chi^2_{2L}$.
% A heuristic is now used to estimate the probability $P(\text{obs}|\text{model})$.
% The probability $P(\text{obs}|\text{model})$ is calculated here as $P(\text{DTW dist} > D|\text{model})$ which can be interpreted as the probability that observations can not agree more with the model than the current set of observations.
% $P(\text{DTW dist} > D|\text{model})$ is calculated as:
\begin{equation}
    \label{eq: heuristic}
    P(\text{\textbf{obs}} |\text{model}_j) = \int_{D_j}^{\infty}\chi^2_{2L_j}(x)dx
\end{equation}
% The probability of the observations $\bs{q_1}\ldots \bs{q_{30}}$ given that the motion follows $\text{model}_j$, i.e. $P(\text{obs}|\text{model}_j)$, is now estimated using $P(D^{+}|\text{model}_j)$.
This is illustrated in figure \ref{fig: chi squared} for a given value of $L_j=3$.
% A low DTW distance $D_j$ results in a high probability and a high DTW distance $D_j$ results in a low probability.
% If all observations lie very close to the model, which is unlikely to happen, we still want to assign a high probability $P(\text{obs}|\text{model}_j)$.
% For this reason, $P(\text{obs}|\text{model}_j)$ is computed using the probability of $D$ or \textit{higher}.

% There is some intuition to this heuristic.
% A low DTW distance indicates a good agreement between sensor data and motion model.
% In figure \ref{fig: chi squared}, if the DTW distance decreases (for the same $\chi^2$ distribution), the shaded area under the $\chi^2$ curve will increase, resulting in higher probability $P(\text{DTW dist} > D|\text{model})$.
% In the extreme case with $D=0$, a perfect match between the sensor data and motion model has been detected and $P(\text{DTW dist} > D|\text{model})=1$.
% Conversely, a very high DTW distance, indicating lower agreement between sensor data and motion model, will lead to a smaller shaded area in figure \ref{fig: chi squared}, which in turn results in a lower $P(\text{DTW dist} > D|\text{model})$.

% Equation \eqref{eq: heuristic} can be applied for every motion model in the database resulting.
However, to calculate the probability that the motion belongs to none of the motion models in the database, i.e. $P(\text{\textbf{obs}}|\text{other})$, we cannot rely on a DTW distance to a motion model (as no model is available).
Therefore, $P(\text{\textbf{obs}}|\text{other})$ is calculated using the following rule of thumb, assuming that the probabilities of not observing \textbf{obs} given model$_j$ are independent:

\begin{equation}
    \label{eq: prob other}
    P(\text{\textbf{obs}}|\text{other}) = \underset{j}{\prod} (1-P(\text{\textbf{obs}}|\text{model}_j))
\end{equation}

The a posteriori model probability of each model (including the ``other'' model), i.e. the probability of a model$_j$ given the observations, is now calculated with Bayes' rule:
\begin{equation}
    \label{eq: a posteriori}
    P(\text{model}_j|\text{\textbf{obs}}) = \frac{P(\text{\textbf{obs}}|\text{model}_j)P(\text{model}_j)}{\sum_{j}P(\text{\textbf{obs}}|\text{model}_j)P(\text{model}_j)}
\end{equation}
These a posteriori probabilities satisfy the condition that $\sum_{j}P(\text{model}_j|\text{\textbf{obs}})=1$.
$P(\text{model}_j)$ corresponds to the prior probability of model$_j$. 
Here, equal priors were chosen such that they disappear from equation \eqref{eq: a posteriori}.
However, these priors can be updated as soon as more usage data of the exoskeleton is available.
Alternatively, the priors can be used as tuning factors to tune the behavior of the controller.

Mei et al. developed a similar approach to use Mahalanobis distances in the Dynamic Time Warping algorithm \cite{Mei2016}.
Furthermore, Gambon et al. converted a Mahalanobis distance to a probability using the $\chi^2$ distribution as well \cite{Gambon2020b}.
To the authors' knowledge, no previous work has combined Mahalanobis Dynamic Time Warping followed by conversion to probabilities using the $\chi^2$ distribution.

\subsubsection{Making Decisions}

\label{sec: making decisions}
Decision theory provides a framework describing how to use the estimated posterior probabilities to make decisions under uncertainty \cite{Daphne2009b}.
To do so, the impact (or utility) of decisions on a user is estimated.
The controller will then take the action that maximizes the expected utility (hence the name Utility Maximizing Controller). 
The Expected Utility ($EU$) of an action $a_l$, given the observations is then calculated as:
\begin{equation}
    \label{eq: expected utility}
    EU(a_l|\text{\textbf{obs}}) = \sum_j U(a_l,\text{model}_j,s_{\text{curr},j})P(\text{model}_j|\text{\textbf{obs}})
\end{equation}
with $P(\text{model}_j|\text{\textbf{obs}})$ given by \eqref{eq: a posteriori} and $U(a_l, \text{model}_j, s_{\text{curr},j})$ the utility of taking action $a_l$ (with $l=1\ldots4$), under the assumption that $\text{model}_j$ corresponds to the actual user motion and that the current progress of the user with respect to model$_j$ is given by $s_{\text{curr},j}$.
The parameter $s_{\text{curr},j}$ used to evaluate equation \eqref{eq: expected utility} is calculated in the onset detection algorithm and is visualized in figure \ref{fig: dtw process}.

The controller will now take the action that maximizes the expected utility:
\begin{equation}
    \label{eq: max eu}
    \underset{a_l}{\text{arg max}} \quad EU(a_l|\text{\textbf{obs}})
\end{equation}

\subsubsection{Calculating Utilities}

\label{sec: calculating utilities}
This section presents an intuitive approach to calculate the utility functions from equation \eqref{eq: expected utility} for each motion$_j$ by exploiting prior knowledge about those motions and about the exoskeleton support mechanism.
The appendix presents a more formal and generally applicable approach to calculate the utility functions using the \textit{value iteration algorithm} known in reinforcement learning.
% Crucial to evaluating equation \eqref{eq: expected utility} are the utility functions $U(a_l, \text{model}_j, s_{\text{curr},j})$.

The utility functions capture the effect of a controller action on a user during a motion, such as locking both sides during a squat motion.
For each motion$_j$ and for each leg separately, the effect of a controller action at progress instance $s_{\text{curr}}$ is calculated as the reduction in human joint torques during the remainder of motion$_j$.
This is referred to as the \textit{total reward} of an action.
The utility functions are subsequently obtained by summation of the total reward functions for both legs.
Because the total reward is calculated using human joint torque data, which cannot be measured at runtime, the total rewards and utility functions are calculated offline and loaded into the controller at start up.

% The effect of action $a_l$ at progress instance $s_{\text{curr}}$ during motion$_j$ is calculated for each leg separately using the reductions in human joint torques during the remainder of the motion, i.e. for $s_n\geq s_{\text{curr}}$ and is referred to as the \textit{total reward}.
% % The effect of $a_l$ at progress $s_{\text{curr}}$ is calculated using the reductions in human joint torques during the remainder of the motion, i.e. for $s_n\geq s_{\text{curr}}$ and is referred to as the \textit{total reward}.
% The total reward functions and the utility functions are calculated offline using a human motion dataset because evaluating the total reward requires human joint torque data, which can not be measured at runtime.
% The resulting utility functions are loaded into the controller at start up.
%  This evaluation requires a dataset of human joint angle and joint torques which is described further on.
% This section presents an approach to calculate the utilities starting from the constraints imposed by the exoskeleton and by using a human joint angle and joint torque dataset.

\textbf{Prior knowledge:}
% If a user is flexing the hips after the support mechanism has been locked, $\alpha$ increases and potential energy is built up in the springs of the support mechanism, resulting in support being provided to a user.
As is clear from figure \ref{fig: torqueDeflection}, the deflection angle $\alpha$, which is the amount of hip flexion after locking the support mechanism, determines the level of support.
% Moreover, no support can be provided if the support mechanism is locked during hip extension.
Thus, to provide sufficient support, the support mechanism should be locked as early as possible during the hip flexion phase of a motion.
For the squat, stoop left and stoop right motions, this means that the controller should lock the support mechanism as soon as possible after a motion has started.

However, because a user's weight is not always equally distributed on both legs, the controller should not necessarily lock both sides.
Support should only be provided to legs that are carrying a user's weight.
Therefore, the controller should lock both sides during squat motions whereas it should only lock the mechanism at the weight supporting side during asymmetric stoop motions.

\begin{figure}
    \centering
    \includegraphics[width=\columnwidth]{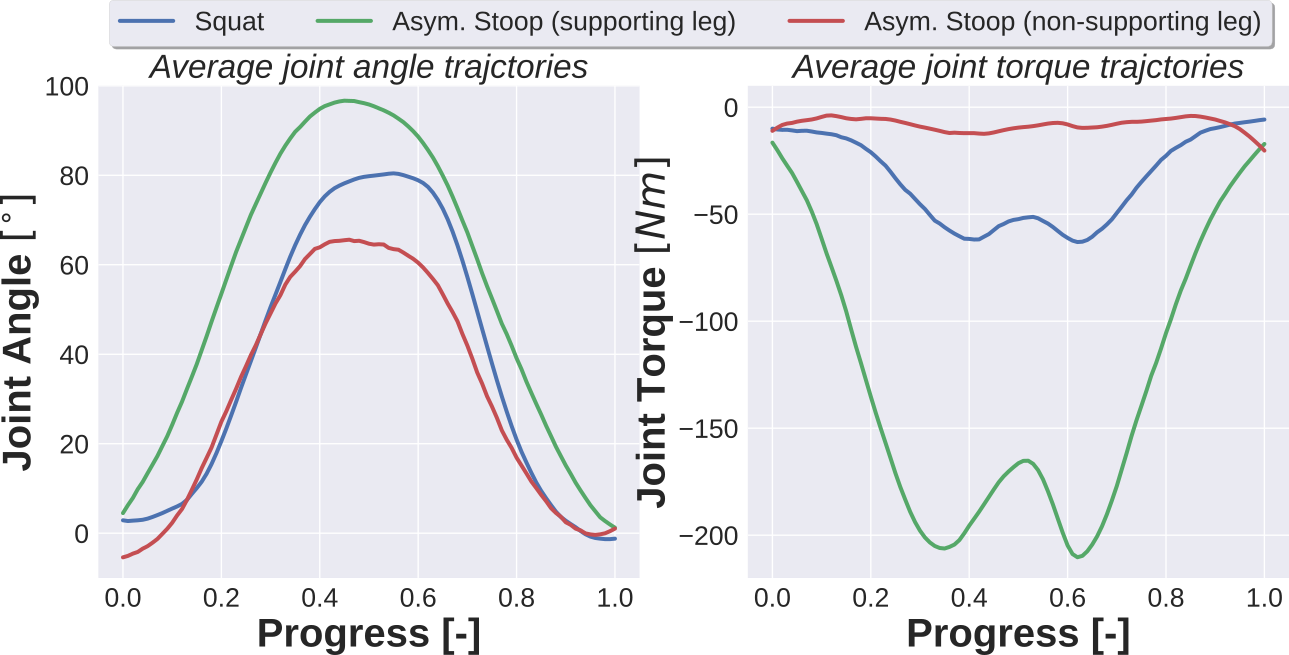}
    \caption{Mean trajectories extracted from data gathered by Van der Have et al. in \cite{VanderHave2019}.}
    \label{fig: squat data}
\end{figure}

Finally, the controller should take into account that deciding to not lock a support mechanism still leaves open the option to lock the support mechanism later on.
In contrast, after locking the support mechanism, the controller cannot decide to unlock the mechanism (as this happens mechanically).
\textbf{Human motion data:}
For each motion$_j$, human joint angle and joint torque data is extracted from a dataset captured by Van der Have et al. \cite{VanderHave2019}.
The human demonstrations are aligned and time-normalized.
Because the human motion data describes the same motions as the PPCA motion models, it is assumed that the same progress variable $s_n$ also describes progress with respect to the human motion data.
From the processed demonstrations, a mean hip joint angle trajectory ($\theta_{\text{hum}}(s_n)$) and hip joint torque trajectory ($\tau_{\text{hum}}(s_n)$) are extracted.
Figure \ref{fig: squat data} visualizes the mean trajectories for the supporting and non-supporting legs during asymmetric stoop motions and for one leg during squat motions (the trajectories for the other leg during squat motions are similar).

\textbf{Total reward:}
The total reward is calculated for each leg separately and for each motion$_j$.
% The total reward for a motion$_j$ is calculated by evaluating the reductions of human joint torques achieved by the exoskeleton from the current progress instance $s_{\text{curr}}$ until the end of the motion.
% This total reward is calculated separately for each leg.
% The utility functions are calculated as the sum of the total rewards for each leg.
To not overload the equations in this section, the subscript $j$ is left out. 
% However, the total reward is calculated for each motion.

The total reward resulting from locking the support mechanism at progress instance $s_{\text{curr}}$ during a motion is calculated using the human $\tau_{\text{hum}}(s_n)$ and exoskeleton joint torques $\tau_{\text{exo}}(\alpha_n)$ and is given by:
\begin{equation}
    \label{eq: total reward lock}
    q(s_{\text{curr}}, \text{lock}) = - \underset{s_n\geq s_{\text{curr}}}{\sum} (\tau_{\text{hum}}(s_n)-\tau_{\text{exo}}(\alpha_n))^{2}
\end{equation}
The square is used to favor reductions of peak human torques over e.g. reductions of average human torques while the minus sign in front ensures that the highest rewards correspond to the highest reductions of human torque.
$\tau_{\text{exo}}(\alpha_n)$ is calculated using the MACCEPA 2.0 actuator model shown in figure \ref{fig: torqueDeflection} with the deflection angle calculated using the mean hip joint angle trajectories $\theta_{\text{hum}}(s_n)$ and equation \eqref{eq: deflection angle}.
 
Calculating the total reward of not locking at $s_{\text{curr}}$ requires an assumption about controller actions at $s_n>s_{\text{curr}}$ as the controller leaves open the option to lock the support mechanism later on in the motion. 
This work assumes that at all instances $s_n>s_{\text{curr}}$, the controller takes actions maximizing the total reward. 

For legs supporting a user's weight, the total reward is maximized by locking the support mechanism as soon as possible.
Therefore, the total reward of not locking (denoted as $\overline{\text{lock}}$) at instance $s_{\text{curr}}$ is
% \thinmuskip=0mu\relax
\medmuskip=0mu\relax
\thickmuskip=0mu\relax
\begin{equation}
    \label{eq: total reward unlock supp}
    \resizebox{0.89\hsize}{!}{$q_{\text{supp.}}(s_{\text{curr}}, \overline{\text{lock}}) = -\tau_{\text{hum}}(s_\text{curr})^2-\!\!\!\!\underset{s_n > s_{\text{curr}}}{\sum}\!\!\!\!(\tau_{\text{hum}}(s_n)-\tau_{\text{exo}}(\alpha_n))^{2}$}
\end{equation}

\begin{figure}
    \centering
    \includegraphics[width=\columnwidth]{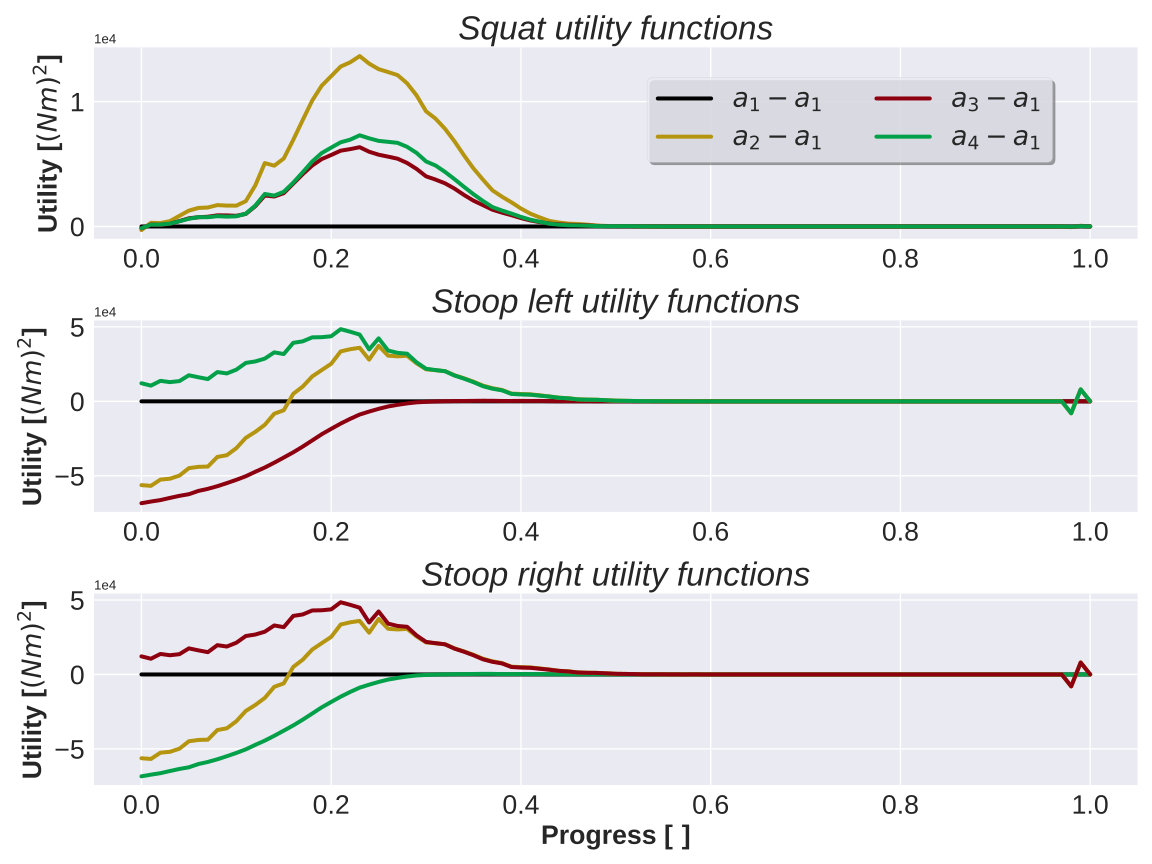}
    \caption{
    Differences between the calculated utility functions for each action $a_l$ during squat motions (\textit{top}), asymmetric stoop left motions (\textit{middle}) and asymmetric stoop right motions (\textit{right}). 
    The curves represent the difference in utility of each action compared to action $a_1$, ``do not lock''.}
    \label{fig: q_functions}
\end{figure}

Conversely, for legs not supporting a user's weight, the total reward is maximized by keeping the support mechanism unlocked:
\begin{equation}
    \label{eq: total reward unlock non supp}
    q_{\text{non-supp.}}(s_{\text{curr}}, \overline{\text{lock}}) = - \underset{s_n\geq s_{\text{curr}}}{\sum}\tau_{\text{hum}}(s_\text{curr})^2 
\end{equation}
% If the controller decides at $s_{\text{curr}}$ to not engage the support mechanism, the opportunity is still available to do so at a later progress instance.
% However, the decision made by a controller also depends on the posterior probabilities which depend on concrete observations.
% Therefore, to evaluate the total reward of not engaging the support mechanism, we must assume what the controller will do later on in a motion.
% Here, we assume that the controller will take the correct actions at all progress instances $s_n>s_{\text{curr}}$.
% Therefore, we assume that for legs supporting weight, the controller locks the support mechanism at the next opportunity while for legs not supporting weight, the controller continues to keep the support mechanism unlocked:
\textbf{Utility functions:}
The utility functions are calculated by summation of the total reward functions;
For example, the utility function for $a_1$ (do not engage) during asymmetric stoop left is given by summation of equation \eqref{eq: total reward unlock supp}, evaluated on data of the left hip joint, and equation \eqref{eq: total reward unlock non supp}, evaluated on data of the right hip joint.
Because of the reward maximizing assumption made in equations \eqref{eq: total reward unlock supp} and \eqref{eq: total reward unlock non supp}, the utility curves for all actions during a given motion lie close to each other.
Therefore, figure \ref{fig: q_functions} visualizes the differences between the utility functions for each action and for the three motions. 
The action $a_1$, ``do not lock'', is used as baseline with which the utilities for the other actions are compared.
Note that for asymmetric stoop motions not locking the support mechanism (i.e. action $a_1$) initially results in higher utility than providing support at the non-supporting leg. 

To evaluate equation \eqref{eq: expected utility}, the utility $U(a_l, \text{other}, s_{\text{curr,}\text{other}})$ should be calculated as well.
However, it is not possible to calculate a total reward for this case because, by definition, this is the reward for unseen motions. 
As a heuristic, and because the exoskeleton is known to counter-act walking motions when the support mechanism is locked, $U(a_l,\text{other},s_{\text{curr,}\text{other}})$ is replaced by the minimal value of $U(a_l, \text{walking}, s_{\text{curr,},\text{walking}})$.

\subsection{Extended Utility Maximizing Controller}
\label{sec: extending intent recognition controller}
The dashed lines in figure \ref{fig: controller_overview} show the extensions made to the Utility Maximizing Controller which resulted in the Extended Utility Maximizing Controller (E-UMC).
These two extensions aimed to further improve and stabilize the behavior of the UMC.

First, a calibration step was added to the startup procedure of the exoskeleton.
After powering on the exoskeleton, users are asked to stand up straight.
The resting position is used to calibrate the sensors to the same starting position.
This compensates for offsets in observations due to imperfect fit of the exoskeleton on a user's waist as well as variations in observations due to repeatedly donning and doffing of the exoskeleton.

Second, a module was added that estimates the left and right hip joint velocities using a Kalman Filter. 
The UMC is only activated if the hip joint velocities exceed a predetermined threshold, i.e. if:
\begin{equation}
    \label{eq: velocity signal}
    \omega_{R,k} \geq 1\text{ rad}/s \text{  AND  } \omega_{L,k} \geq 1\text{ rad}/s.
\end{equation}

\subsection{Passive Exoskeleton Controller}
\label{sec: velocity trigger}
To benchmark the UMC and E-UMC on the hip exoskeleton, a controller emulating a passive exoskeleton was developed (Passive Exoskeleton Controller, PEC) as well.
This also allowed for comparison of all the three controllers independently of the exoskeleton hardware.

% A controller emulating a passive exoskeleton is developed to benchmark the proposed controller and to avoid the need for a second, passive exoskeleton. 
% Hence, the same exoskeleton prototype can be used in the evaluation such that results are not influenced by the hardware design.

In a passive exoskeleton the support mechanism is always locked, both when bending forward and when walking around.
Hence, users often have to work against the support mechanism (as concluded by \cite{Naf2018a}).
However, Kazerooni et al. discuss the \textit{passive torque generator} implemented on the backX exoskeleton which allows unhindered walking but generates assistance during squat or stoop motions \cite{Kazerooni2019}.

Therefore, we chose to model a passive exoskeleton with such a passive torque generator.
This simple controller locks the support mechanism if the left and right hip joint velocities increase above a predefined threshold, i.e. if the conditions from equation \eqref{eq: velocity signal} are satisfied.
% Note that this immediately exposes an additional benefit of quasi-passive devices: the exoskeleton hardware no longer imposes the support strategy.

\subsection{Practical Implementation}
\label{sec: practical implementation}
The above controllers were implemented in C++ in the OROCOS real-time component system \cite{Bruyninckx2001}.
The Utility Maximizing Controller runs at 10Hz.
The Passive Exoskeleton Controller runs at 100Hz since it has a lower computational load.
The Extended Utility Maximizing Controller runs at 100 Hz as long as the velocity thresholds are not exceeded, otherwise it runs at 10Hz.
Both controllers interact with a low-level OROCOS controller running at 100Hz.
The low-level controller communicates with the exoskeleton through the EtherCAT protocol \cite{EtherCAT}.
All code is executed on a Dell Latitude 5590 laptop with 16Gb RAM and an Intel Core i7 processor running Ubuntu 20.04 LTS.

%%% Local Variables:
%%% mode: latex
%%% TeX-master: "../root"
%%% End:

%% file: sections/experiments.tex
\section{User study set-up}
\label{sec: user study}

A user study was conducted to evaluate the proposed controllers.

A first objective of the user study was to evaluate the performance of the controllers during lifting motions executed by different participants.
% A first objective of the user study is to evaluate the performance of the controller during lifting motions for various participants.
% The controller should be able to deal with the variation displayed by different participants executing the same motions. 
The motions of a participant should reliably and intuitively trigger the exoskeleton to provide support during lifting motions.
In turn, the controllers should generate appropriate joint torques to support a user during lifting motions.
In particular, the exoskeleton should generate symmetric torques to the hip joints during squat motions and should generate asymmetric torques during asymmetric lifting motions.

A second objective was to evaluate the robustness of the controller during non-lifting motions as workers in industry are likely to walk around their work station.
Therefore, the exoskeleton should not hinder users during motions such as walking, climbing stairs or descending stairs.

The final objective was to qualitatively investigate user preferences towards the different controllers. 
In case of a clear user preference, it could be taken into account in future versions of the exoskeleton.
In this study, qualitative user preferences were investigated between the UMC and PEC.

The remainder of this section discusses the study design, the data acquisition and processing and the evaluation metrics.
Section \ref{sec: results} presents the results while section \ref{sec: discussion} discusses the results.

\subsection{Study Design}
The user study was completed in two phases.
In a first, data collection phase, exoskeleton data was recorded to create PPCA motion models. 
% The procedure to obtain human joint angle and joint torque data required for the utility database in figure \ref{fig: controller_overview} requires human motion demonstrations in a gait laboratory and is described in \cite{VanderHave2019}.
In a second, evaluation phase, all three controllers were evaluated in a series of lifting and walking motions.
Throughout the evaluation phase, the set of motion models remained unchanged.

\subsubsection{Data Collection Phase}
% Ages for phase 1 were: 27, 30, 22, 26, 25, 26
Six participants (including one of the authors, all male, average age: $26 \pm 2.6$) were asked to perform squat, stoop left and stoop right motions while wearing the exoskeleton. 
These motions were recorded with the exoskeleton and used to extract three PPCA motion models.
Prior to the start of the motion sequences, participants received detailed information about the goal and setup of the study and provided written informed consent.
The study was approved by the local ethics committee (Universitair Ziekenhuis Leuven, file number: S61611).
In total, participants performed ten squat motions, five asymmetric stoop left motions and five asymmetric stoop right motions.
Throughout these motions, participants were lifting a box of ten kilograms.
The width of the exoskeleton was adjusted to ensure a proper fit.
Before starting the sequence of motions, participants were allowed to move around and familiarize themselves with the exoskeleton.

In the squat sequence, participants were asked to pick up the box from the ground, stand up straight for one second and subsequently put the box back on the ground.
This was repeated five times.
% In the squat motions, the box was placed in front of the participants. 
% Participants were asked to put their feet at shoulder width with the toes pointing forward.
% Next, participants had to pick up the box by flexing both hips and knees while keeping the back straight but were free to execute the squat motions at self-selected speeds.
% During the asymmetric stoop motions, participants were asked to put feet at shoulder width at an angle of 90 degrees.
% The box was put in front of one foot. 
% Participants had to pick up the box, come back up to resting position and put the box down in front of the other foot.
In the asymmetric stoop sequence, participants were asked to put their feet at an angle of 90 degrees with respect to each other.
Participants had to pick up the box in front of one foot, stand up straight for one second and put the box down in front of the other foot.
Again, this was repeated five times.
Participants could execute all motions at self-selected speeds.
Due to the low weight of the box, it was assumed that squat and stoop motions with and without holding a box are similar in execution.

\subsubsection{Evaluation Phase}
% Ages for phase 2 were: 33,26,23,27,23,27,24,22
% Ages for EIRC evaluation were: 26,23,24
In the evaluation phase, the UMC and PEC were evaluated with eight participants (7 male, 1 female, average age: $25.6\pm3.3$). 
Three out of the eight participants evaluated the E-UMC as well (3 male, average age: $24 \pm1.5$).
All participants received detailed information about the goal and setup of the study and provided written informed consent.
This part of the study was approved by the KU Leuven Social and Societal Ethics Committee (file number: G-2022-5430).
Only one participant participated in both phases of the user study.

% Participants were asked to perform the same lifting motions as in Phase I of the user study.
In this phase, participants executed the same sequence of lifting motions as in the data collection phase.
To limit fatigue for the participants, the weight of the box was limited to five kilograms.
In addition to the lifting motions, participants performed a walking and stair walking trial.
In the walking trial, participants had to walk 10 meters in a straight line, turn around and walk back.
In the stair walking trial, participants had to walk up eleven stairs, turn around on a landing and walk down eleven stairs.

Participants were allowed to pause in between the lifting motions and after each sequence of lifting, walking and stair walking motions.
Although the locking mechanism was explained to participants, they did not know any details about the controllers before each motion sequence.
Across all participants, the order in which the UMC and PEC were evaluated was randomized. 
The E-UMC was evaluated after the evaluations with the UMC and PEC were completed.

\subsection{Data acquisition and processing}
\subsubsection{Data Collection Phase}
Encoders on the exoskeleton registered hip flexion angles during the squat and stoop sequences.
% Throughout the sequence of squat and stoop motions, the encoders on the exoskeleton were continuously registering hip flexion angles.
These sequences were first manually segmented into individual lifting motions.
However, only the first half of each lifting motion corresponding to hip flexion was stored.
In total, the segmented dataset consisted of 60 squat demonstrations, 30 stoop left demonstrations and 30 stoop right demonstrations. 
These demonstrations were processed according to section \ref{sec: controller overview} and PPCA motion models were extracted.
Figure \ref{fig: motion models} visualizes the resulting motion models which were stored in the motion model database of figure \ref{fig: controller_overview}. 

\subsubsection{Evaluation Phase}
Apart from the hip flexion angles, some controller variables were registered as well.
% Apart from the hip flexion angles, intermediate results from the onset detection algorithm and the decision algorithm were also recorded to simplify a posteriori analysis.
For each motion model, the DTW distance, progress estimates, probability estimates and warping function length were calculated and registered as well as the control signal sent from the high-level controller to the low-level controller. 
% The control signal sent from the high level controller to the low level controller is stored as well. 
All lifting motion sequences were manually segmented into (complete) individual lifting motions. 

% Apart from the quantitative data, some qualitative data was obtained as well.
To evaluate user preferences, qualitative data was collected.
After completion of a sequence of motions with the UMC or PEC, participants were asked to fill in two questionnaires.
The first questionnaire was the System Usability Scale (denoted as SUS) \cite{Brook1995}.
Responses on the SUS questionnaire were processed according to the guidelines in \cite{Brook1995}.
The second questionnaire aims to evaluate user acceptance of exoskeletons and was proposed by Maurice et al. in \cite{Maurice2020}.
It is denoted here as Exoskeleton Usability Scale (EUS).
After completing the study with the UMC and PEC, participants were explicitly asked whether they had a preference for either the UMC or PEC.
The three participants that additionally evaluated the E-UMC, did not have to fill in the questionnaires a third time. 

\subsection{Evaluation}
The recorded quantitative data served as a basis to compare performance of the three controllers.
First, intent estimation performance was evaluated by counting the number of recognized motions.
For the purpose of analysis, a motion was assumed to be correctly recognized if the detection algorithm assigned a probability $P(\text{model}|\text{obs})\geq 0.5$ to it during the first half of a complete lifting motion.

Second, the number of correct decisions, taken in the first half of a lifting motion, was counted as well. 
A correct decision was to lock both sides during a squat motion, lock left side during stoop left motions and lock right side during stoop right motions. 
Note that, according to the utility functions, it is optimal to lock both sides near the end of the hip flexion motion in asymmetric lifting motions.
However, the first decision during an asymmetric stoop motion should always be to engage the supporting side. 

Third, the number of false positive lock signals was counted to evaluate the robustness of the controller during non-lifting motions. 
% This provides a measure of the robustness of the evaluated controller to non-lifting (walking and stair walking) motions.
Finally, the generated exoskeleton torques were calculated a posteriori using the control signals sent to the low-level controller and the spring model.
Since the exoskeleton should have locked both support mechanisms during squat motions, symmetric exoskeleton torques were to be expected during squat motions.
In contrast, asymmetric exoskeleton torques were to be expected during asymmetric lifting motions with the UMC or E-UMC.

In the qualitative analysis, the scores with the SUS and EUS were compared between the UMC and PEC and the highest score was assumed to indicate user preference.

%% file: sections/results.tex
\section{Results}
\label{sec: results}
This section presents and discusses the results of the user study.
% \subsection{Experiment 1}
% \label{sec: experiment 1 results}
% The main result for the first round of experiments are the recorded demonstrations from the five participants. 
% These motions were a posteriori segmented and resampled such that a motion model could be extracted using the methods from section \ref{sec: motion models}.
% In total, 10 squat and stoop demonstrations from each participant as well as from one of the authors were used. 
% Thus, 60 demonstrations were used to generate the squat motion model and 30 demonstrations were used for each of the asymmetric stoop motion models.
% The resulting motion models are visualized in figure \ref{fig: motion models}.
\begin{figure}
    \centering
    \includegraphics[width=\columnwidth]{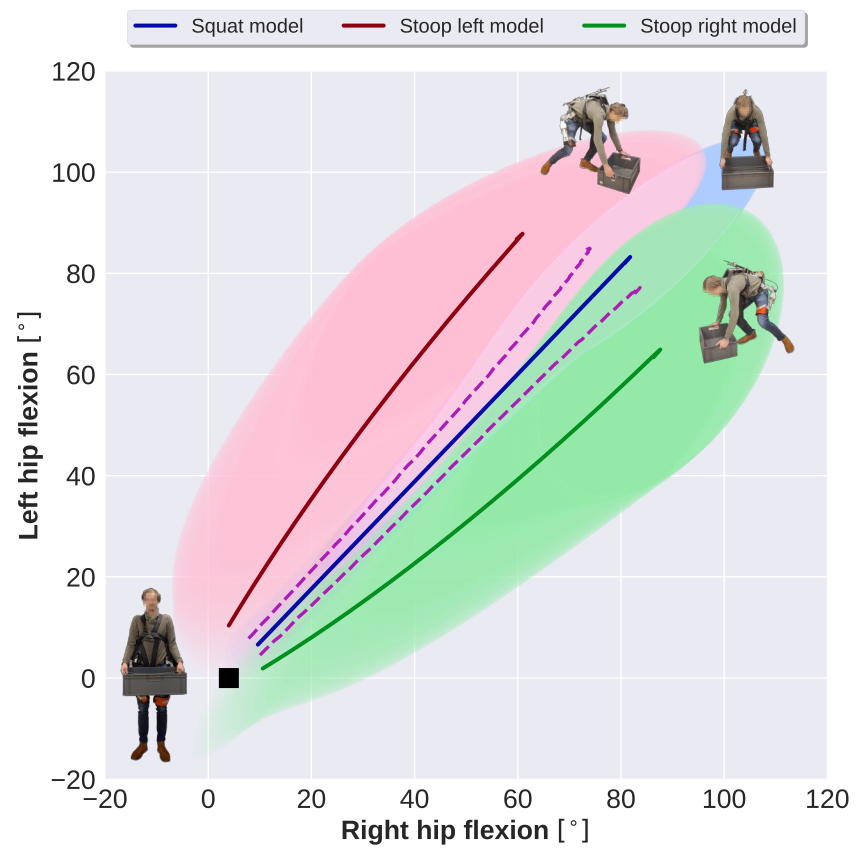}
    \caption{This figure shows the different motion models stored in the motion model database. 
    The mean of each model is shown as well as the 95\% confidence interval around the mean. 
    The magenta dotted line connects points of equal Mahalanobis distance between the squat and stoop left model as well as between the squat and stoop right model.
    The models start in standing position (lower left corner of the figure) and end in fully flexed position (top right corner of the figure). 
    The marker in the lower left corner indicates the resting position after calibration for the E-UMC.}
    \label{fig: motion models}
\end{figure}

\subsection{Qualitative Results}
Table \ref{tab: questionnaire results} presents the results of the SUS and EUS questionnaires as well as the preferences indicated by participants when explicitly asked about it.
For each participant and each questionnaire, the value in bold highlights the highest score.
The SUS scores have a maximum value of 100, while the EUS scores have a maximum value of 10. 

The SUS and EUS scores vary a lot between participants rendering inter-participant comparisons difficult.
In contrast, the scores given by a participant in each questionnaire to both controllers are very close. 
Note that preferences in the SUS questionnaire seem not to correspond with preferences in the EUS questionnaire as well as the explicit preferences indicated in the bottom row of the table.
Only participant 4 consistently indicated a preference for the PEC. 

% From this table, no clear conclusion can be drawn about user preferences towards the UMC and PEC.
% Although the next subsection will show that significantly different support torques were generated using the three evaluated controllers, participants seemed not to experience large differences. 
  
\begingroup
\setlength{\tabcolsep}{5pt}
\begin{table}[t]
    \centering
    \caption{Questionnaire Results for PEC and UMC}
    \label{tab: questionnaire results}
    \begin{tabular}{llcccccccc}
        \toprule
        & & \multicolumn{8}{c}{Participant} \\ 
        & & 1 & 2 & 3 & 4 & 5 & 6 & 7 & 8 \\ 
        \midrule
        \multirow{2}{*}{SUS} & PEC & 72.5 & \textbf{97.5} & 72.5 & \textbf{82.5} & 62.5 & 92.5 & \textbf{80} & \textbf{75} \\
                             & UMC & \textbf{75} & 95 & \textbf{80} & 80 & \textbf{70} & \textbf{95} & 77.5 & 70 \\ 
        \midrule
        \multirow{2}{*}{EUS} & PEC & 6.35 & 7.8 & 7.1 & \textbf{8.4} & \textbf{6.45} & \textbf{8.4} & 4.95 & \textbf{6.6} \\
                             & UMC & 6.35 & \textbf{7.95} & \textbf{7.95} & 7.95 & 6.25 & 8.15 & \textbf{5.65} & 5.95 \\ 
        \midrule
        \multicolumn{2}{c}{Preference} & / & / & / & PEC & PEC & PEC & UMC & /  \\
        \bottomrule
    \end{tabular}
\end{table}
\endgroup

\subsection{Quantitative Results}
\subsubsection{Intent estimation and decision performance}
Table \ref{tab: recognition and decision performance} shows the number of correctly recognized motions as well as the number of correct decisions for squat (SQ) and asymmetric stoop (AS) motions (both with a maximum of 10 per participant and per motion type).
The table also presents the results for the three participants (participants 2, 3 and 7) that evaluated the E-UMC.
The UMC controller is able to recognize ongoing motions, and make a correct decision but did not seem to work properly for participant 3
Additionally, the decision step in the UMC had relatively low performance during the stoop motions of participant 6 and 7. 
For all other participants, the UMC controller worked as intended.
The improvements of the E-UMC with respect to the UMC are clearly visible for participants 3 and 7 in table \ref{tab: recognition and decision performance}.
% The additional calibration of the resting position in the E-UMC facilitates the intent estimation step of the E-UMC. 
% For participants 3 and 7, the improvement is clearly visible in table \ref{tab: recognition and decision performance}.

The PEC, which uses only the instantaneous joint velocities, engaged both support mechanisms during all demonstrations for all participants except for participant 7.
For this participant, the PEC did not support three squat motions and one asymmetric stoop motion. 

\begingroup
\setlength{\tabcolsep}{5pt}
\begin{table}
    \centering
    \caption{Intent estimation and Decision Performance for (E-)UMC}
    \label{tab: recognition and decision performance}
    \begin{tabular}{llcccccccc}
        \toprule
        \multicolumn{2}{c}{\multirow{2}{*}{UMC/E-UMC}} & \multicolumn{8}{c}{Participant} \\ 
        \multicolumn{2}{c}{}                              & 1 & 2 & 3 & 4 & 5 & 6 & 7 & 8 \\
        \midrule
        Intent                       & SQ & 7 & 10/10 & 8/10 & 10 & 10 & 10 & 10/10 & 10 \\
        % \multirow{2}{*}{Recognition} & SQ & 7 & 10/10 & 8/10 & 10 & 10 & 10 & 10/10 & 10 \\
        Estimation                   & AS & 10 & 10/10 & 5/10 & 10 & 10 & 10 & 6/10 & 10 \\ 
        \midrule
        \multirow{2}{*}{Decision}    & SQ & 7 & 10/10& 6/10 & 10 & 10 & 10 & 10/10 & 10 \\
                                     & AS & 10 & 10/10 & 5/10 & 10 & 10 & 5 & 7/10 & 10 \\ 
        \bottomrule
        
    \end{tabular}
\end{table}
\endgroup

\subsubsection{False positives}
Table \ref{tab: false positives} lists all the false positive lock signals generated by each controller during the walking (W) and stair walking (SW) trials.
As can be seen, the PEC did not generate any false positive lock signal.
In contrast, the UMC did generate quite a few false positive lock signals.
During walking trials, the UMC generated false positive signals for participants 3, 5 and 7.
During all stair walking trials, the UMC generated false positives signals.
% Closer inspection of the estimated motion model probabilities learns that the intent estimation algorithm is not able to discriminate between stair walking and asymmetric stoop lifting.
% Note that participants did not experience high hip torques during walking and stair walking trials since the deflection angle $\alpha$ always remained limited.
In contrast to the UMC, the E-UMC did not generate any false positive lock signals. 
Note that participants did not experience high hip torques during walking and stair walking trials since the deflection angle $\alpha$ always remained small.
% The velocity trigger signal used in the E-UMC is effective at preventing false positive lock signals and improves the robustness of the controller.

\begingroup
\setlength{\tabcolsep}{5pt}
\begin{table}
    \centering
    \caption{Number of false positive lock signals}
    \label{tab: false positives}
    \begin{tabular}{llcccccccc}
        \toprule
        & & \multicolumn{8}{c}{Participant} \\ 
        \multicolumn{2}{c}{}                              & 1 & 2 & 3 & 4 & 5 & 6 & 7 & 8 \\
        \midrule
        \multirow{2}{*}{PEC} & W & 0 & 0 & 0 & 0 & 0 & 0 & 0 & 0 \\
                             & SW & 0 & 0 & 0 & 0 & 0 & 0 & 0 & 0 \\ 
        \midrule
        \multirow{2}{*}{UMC}    & W & 0 & 0 & 7 & 0 & 1 & 0 & 14 & 0 \\
                                & SW & 6 & 16 & 15 & 1 & 15 & 19 & 14 & 20 \\ 
        \midrule
        \multirow{2}{*}{E-UMC}    & W & / &  0 & 0 & / & / & / & 0 & / \\
                                 & SW & / & 0 & 0 & / & / & / & 0 & / \\                                      
        \bottomrule
        
    \end{tabular}
\end{table}
\endgroup

\subsubsection{Exoskeleton torques}
\begin{figure}
    \centering
    \includegraphics[width=\columnwidth]{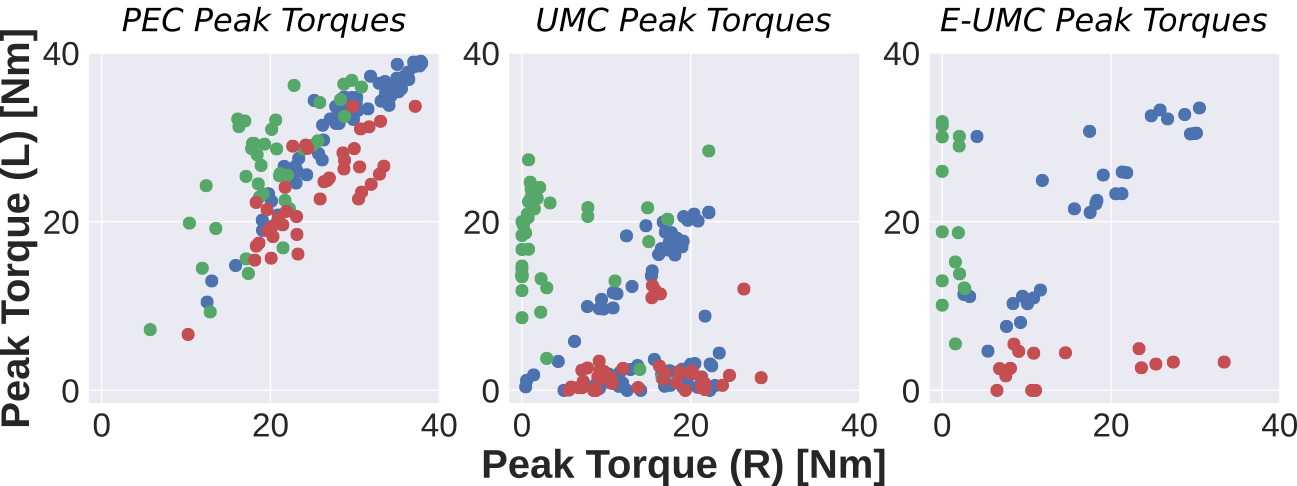}
    \caption{Visualization of the peak of the left (vertical) and right (horizontal) exoskeleton torques during the user study for squat (\textbf{blue}), stoop left (\textbf{green}) and stoop right (\textbf{red}) trials.
    Each marker represents the peak right and left torques during a lifting motion.
    \textit{Left:} Exoskeleton torques with Passive exoskeleton controller (PEC)
    \textit{Middle:} Exoskeleton torques with Utility Maximizing controller (UMC).
    \textit{Right:} Exoskeleton torques with the Extended Utility Maximizing controller (E-UMC).
    }
    \label{fig: peak torques}
\end{figure}
Figure \ref{fig: peak torques} shows the left and right peak torques generated by the exoskeleton during each recorded lifting motion with the three controllers. 
Several aspects from figure \ref{fig: peak torques} should be mentioned.

First, the PEC generates almost identical assistance torques during all motions.
This behavior is what can be expected from a passive exoskeleton.
Peak torque values during stoop left (green) and stoop right (red) show slight asymmetries due to the asymmetries in hip flexion angles during stoop left and stoop right motions.

In contrast, the UMC and E-UMC are able to generate distinctively different support torques during the three lifting motions.
% This is one of the main advantages of using the intent estimation algorithm.
During stoop left motions, only the left hip joint is supported. 
Accordingly, the green markers are grouped near the vertical axis.
Similarly, the red markers (representing peak torques during stoop right motions) are grouped near the horizontal axis.
Peak torques during squat motions lie on the diagonal.
% Because of the differences in weight distribution on the legs during the three lifting motions, the UMC and E-UMC provide peak torques that more closely match human peak torques.

Second, the PEC generates higher peak torques than the UMC during all lifting motions. 
The peak torques generated with the E-UMC vary across subjects, but the controller is also able to generate higher peak torques than the UMC. 
% because of the simpler control structure of the PEC, which only takes instantaneous joint angle velocities into account, it generally reacts faster than the UMC and E-UMC.
% Consequently, the PEC generates higher exoskeleton joint torques.
% Second, the PEC generates considerably higher support torques than the UMC and to a lesser extent than the E-UMC as well.
% This is due to the simpler control structure of the PEC. 
% This controller only takes instantaneous joint angle velocities into account.
% In contrast, both the UMC and E-UMC need to process a horizon of 30 joint angle samples.
% Due to the higher computational load and because the UMC and E-UMC will only lock the support mechanism if the controller is sufficiently certain about the estimated intent, the UM and E-UM will generally react slower to motions than the PEC.
% Consequently, spring extension in the support mechanism is lower, which results in lower exoskeleton torques.
%
% However, this can be easily compensated for by using stiffer springs or increasing the pretension of the support mechanism when using the UMC or E-UMC.

Third, for some participants, peak squat torques generated with the UMC are not grouped along the diagonal but are grouped along the horizontal axis as well. 
% This is surprising as table \ref{tab: recognition and decision performance} indicates that the intent estimation and decision performance of the controller was high for squat motions.
% In reality, the intent estimation algorithm could often not sufficiently distinguish the initial part of a squat motion from a stoop right motion.
% Since the decision algorithm takes safe choices, the controller initially only locked the right support mechanism.
% After observing an additional part of the squat motion, the probability of the motion being a stoop right motion decreased and the controller locked the left support mechanism as well.
% Because of this delayed locking of the left support mechanism, large torque asymmetries occurred during squat motions.
This behavior is no longer present with the E-UMC which generates symmetric peak support torques. % visualized in figure \ref{fig: peak torques}.
% Additionally, the E-UMC generates higher peak support torques than the UMC as squat motions are detected earlier.

% To avoid asymmetric support during squat motions, the detection algorithm has to recognize squat motions early and with a sufficiently high certainty to lock both support mechanisms at the same time.

\begin{figure*}
    \centering
    \includegraphics[width=\textwidth]{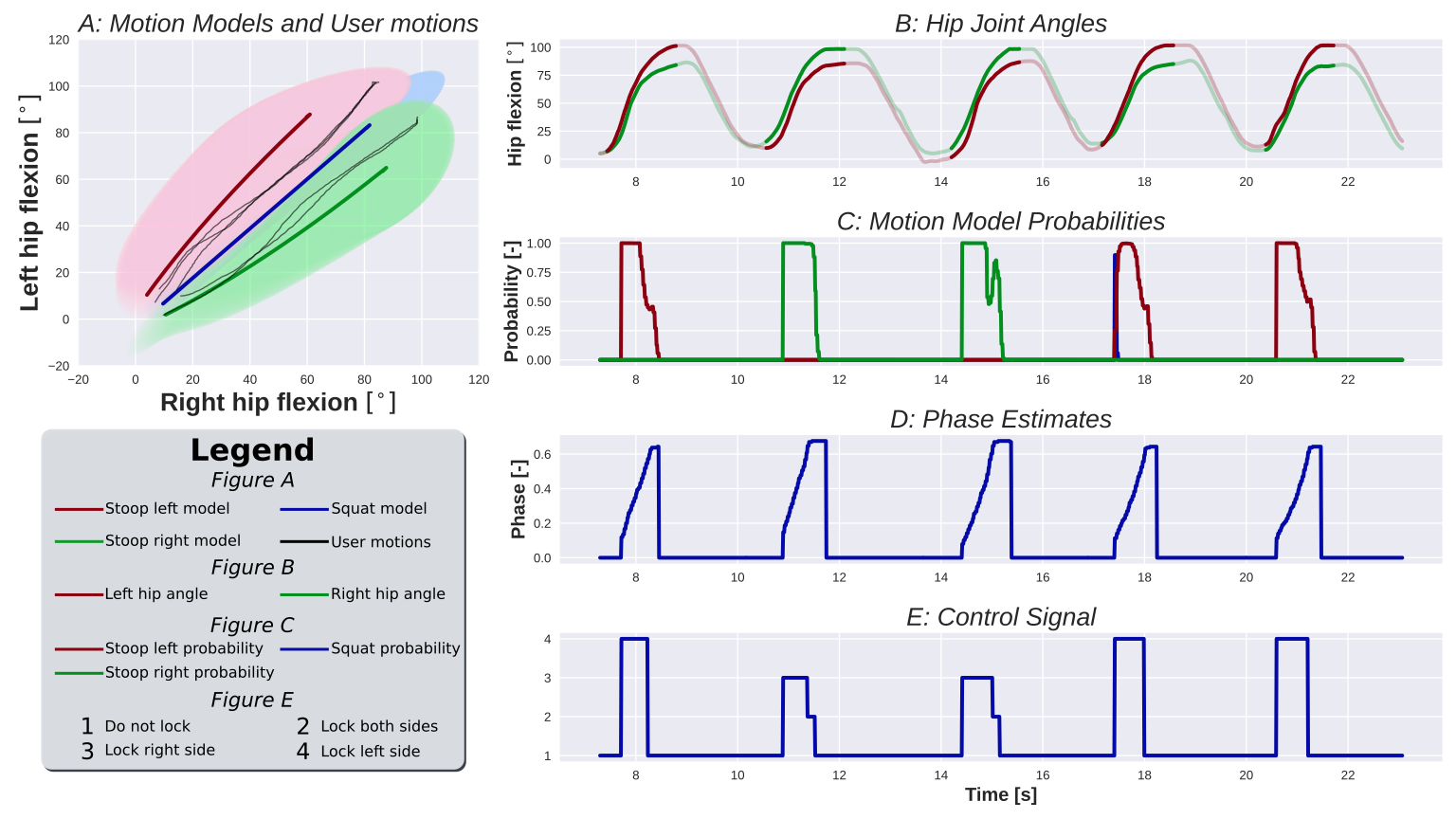}
    \caption{This figure provides an overview of the exoskeleton controller output. 
    The figure visualizes the first five demonstrations of the asymmetric stoop lifting trial of Participant 3 with the E-UMC. 
    (A) shows the motion models used to perform onset detection and intent estimation.
    (B) shows the recorded hip joint angles as a function of time. 
    The hip-flexion part of each lifting trial is non-transparent. 
    These parts are also plotted by black lines on top of the motion models in figure (A).
    (C) shows the estimated motion model probabilities calculated using the onset detection algorithm and the $\chi^2$ distribution.
    (D) shows the progress estimates ($s_{\text{curr}}$) calculated by onset detection algorithm. 
    The figure plots the progress estimate belonging to the motion model that is most probable.
    In reality, a progress estimate is generated for each recognized motion model.
    (E) shows the actual control signal sent to the low-level controller. 
    The legend clarifies the meaning of these control signals.}
    \label{fig: exoskeleton output}
\end{figure*}

To conclude this section, the exoskeleton output is shown in figure \ref{fig: exoskeleton output} for the first five asymmetric stoop lifting demonstrations of Participant 3 with the E-UMC. 
The hip-flexion part of each lifting motion is plotted on top of the motion models. 
The motion model probabilities, progress estimates (given by $s_{\text{curr}}$ in figure \ref{fig: dtw process}) and control signals are plotted. 
Notice how the fourth motion generates a peak in squat probability. 
The corresponding motion indeed starts in the `squat' region of figure \ref{fig: exoskeleton output} (A).
Finally, remark that the progress never reaches $s_{\text{curr}}=1$ as the onset detection algorithm stops running as soon as either $\omega_{R,k}<1 \text{rad}/s$ or $\omega_{L,k}<1 \text{rad}/s$.

%%% Local Variables:
%%% mode: latex
%%% TeX-master: "../root"
%%% End:

%% file: sections/discussion.tex
\section{General discussion and conclusion}
\label{sec: discussion}
The proposed UMC and (E-UMC) controllers assume that the exoskeleton joint torques do not significantly influence a user's motion. 
This assumption allows to decouple the intent estimation step from the decision step such that the utility functions can be calculated offline using human motion data.
This is important as the utility functions rely on human joint torque data which cannot be measured on a factory work floor.
At runtime, these utility functions are evaluated efficiently.
Although prior knowledge about the lifting motions was exploited to calculate the utility functions in section \ref{sec: calculating utilities}, a general approach to calculate the utilities without relying on prior knowledge was formulated in the appendix.

Because of their reliance on human joint angle and joint torque data, the utility functions should be calculated once and reused across exoskeleton users.
On the other hand, the PPCA motion models only rely on user demonstrations recorded with the exoskeleton.
Consequently, the PPCA motion model database can be easily updated with user-specific demonstrations to improve the intent estimation performance.

From table \ref{tab: questionnaire results} no clear conclusion can be drawn about user preferences towards the UMC and PEC.
Although significantly different support torques were generated using the three evaluated controllers, participants seemed not to experience large differences.
Evaluating the exoskeleton over a longer period of time during deployment on the factory floor could result in more clear user preferences.

During stair walking trials, the UMC generates a lot of false positives. 
Closer inspection of the estimated motion model probabilities learns that the intent estimation algorithm in the UMC is not able to discriminate between stair walking and asymmetric stoop lifting.
In contrast, the velocity trigger signal used in the E-UMC is effective at preventing false positive lock signals and improves the robustness of the controller.

Because the PEC only takes into account instantaneous velocities, it reacts faster than the UMC or E-UMC which results in higher peak torques.
However, this can be compensated for by using stiffer springs or by increasing the pretension of the support mechanism when using the UMC or E-UMC.

A crucial advantage of both the UMC and the E-UMC over the PEC is the ability to generate asymmetric support torques during asymmetric lifting motions.
Because of the differences in weight distribution on the legs during the three lifting motions, the UMC and E-UMC provide peak torques that more closely match human peak torques.

Unfortunately, for some participants the UMC generated asymmetric peak torques during squat motions.
This is surprising as table \ref{tab: recognition and decision performance} indicates that the intent estimation and decision performance of the controller was high for squat motions.
In reality, the intent estimation algorithm could often not sufficiently distinguish the initial part of a squat motion from a stoop right motion.
Since the decision algorithm takes safe choices, the controller initially only locked the right support mechanism.
After observing an additional part of the squat motion, the probability of the motion being a stoop right motion decreased and the controller locked the left support mechanism as well.
Because of this delayed locking of the left support mechanism, large torque asymmetries occurred during squat motions.

The calibration step in the E-UMC ensures that the intent estimation algorithms can detect the correct motions with sufficiently high certainty such that both support mechanisms are locked at the same time during squat motions.
Additionally, the E-UMC recognizes motions earlier which results in higher peak torques.
The improvement in intent estimation and decision performance is also clearly visible in table \ref{tab: recognition and decision performance} for participants 3 and 7.

This improvement with the E-UMC shows the importance of calibrating the sensors on the exoskeleton.
This has consequences for the design of the exoskeleton hardware as well. 
In particular, movement of the exoskeleton on the hips during operation should be reduced to limit variation in sensor readings when standing in e.g. resting position.
Additionally, these movements of the exoskeleton result in changing contact points of the leg cuffs and hip interface and can sometimes be painful to users.

One assumption made in the (E-)UMC requires further discussion. 
The priors of all motion models are chosen to be equal in equation \eqref{eq: a posteriori}.
Although it is reasonable that the prior probability of each motion model in the database is equal, it is not as straightforward to assume that the probability of the category ``other'' motions is equal as well.
However, assuming equal priors can be a first starting point for the controller.
As soon as data is available on typical usage patterns, these priors can be updated.
Additionally, monitoring usage patterns over a longer time horizon can generate useful insights in a user's ability to learn how to reliably trigger locking of the support mechanisms.

% Second, the a priori calculation of the utility functions assumes that the human joint angle trajectories are not significantly influenced by the exoskeleton.
% Under this assumption it is possible to first calculate the impact of the exoskeleton during all motions and combine this at runtime with estimates of the probability of each motion.
% If this assumption does not hold, then the impact of the exoskeleton can not be evaluated a priori as it is unclear how joint angle trajectories are affected by the exoskeleton.
% Future work should evaluate whether this assumption holds.

Currently, it is unclear to what extent the exoskeleton is able to reduce muscle activity as measured by electromyography.
Therefore, future work can include a user study in a dedicated gait lab. 
Results from this study should be used to determine the optimal peak torque values that the exoskeleton should provide.
Providing too much assistance could cause a user to lose balance and should be avoided. 

This paper presented controllers for a quasi-passive hip exoskeleton.
The Extended-Utility Maximizing Controller is able to distinguish between three lifting motions and allows users to walk and stair walk.
The methods implemented by the E-UMC are applicable on various motion types or quasi-passive exoskeletons as long as data is available to create the motion model database and utility database.

%% file: sections/appendix.tex
\section*{General approach to calculating utilities}
The utility functions can also be calculated as the rewards received by the controller when simulating its behavior on human joint angle and joint torque data for varying $a_l$, $\text{model}_j$ and $s_{\text{curr},j}$.
The simulation uses the \textit{value iteration algorithm} and does not rely on knowledge of the optimal support strategy in each motion$_j$.
The simulations are run offline for each motion$_j$ using the same human motion dataset introduced in section \ref{sec: calculating utilities}, and separately for the left and right side of the hip exoskeleton.
The results for left and right side are then summed to obtain the complete utility functions.
Details of the value iteration algorithm can be found in \cite{Sutton2018}.

Apart from human motion data, the value iteration algorithm requires a description of the states of the exoskeleton, possible controller actions, a transition model describing how actions influence the state and an instantaneous reward model capturing the effect of the exoskeleton on a user at a specific progress instance $s_n$.
To not overload the equations in this subsection, the index $j$, referring to a particular model, is left out.

\textbf{Exoskeleton state and controller actions:}
The state vector for one side of the exoskeleton at a given progress instance $n$ is written as:

\begin{equation}
    \label{eq: state vector}
    x_n = \begin{cases}
                l_n & \text{locking status of mechanism (0 or 1)} \\
                \theta_{\text{lock},n} & \text{hip flexion angle at instance of locking}\\
                s_n & \text{progress value}
        \end{cases} % '.' to close \{ command above
\end{equation}
The progress value indicates the progress along motion $\theta_{\text{hum}}(s_n)$.
The actions with which the controller can change the state of one side of the exoskeleton are:
\begin{equation}
    u_n = \begin{cases}
                0 & \text{do not change locking status}\\
                1 & \text{engage lock.}
        \end{cases}
\end{equation}

\textbf{Transition model:}
The transitions between states due to the controller actions are:

\begin{equation}
    l_{n+1}= \begin{cases}
    1 & \hspace*{0mm} \text{if $l_n=0$ and $u_n=1$} \\
    0 & \hspace*{0mm} \text{if $l_n=1$ and $\alpha_n < 0$} \label{eq: lock transition} \\
    l_{n} & \hspace*{0mm} \text{otherwise} 
    \end{cases}
\end{equation}

\begin{equation}
    \theta_{\text{lock},n+1}= \begin{cases}
    0  & \hspace*{0mm} \mbox{if $l_n=0$ and $u_n=0$} \\
    \theta_{\text{hum}}(s_n) & \hspace*{0mm} \mbox{if $l_n=0$ and $u_n=1$} \label{eq: theta transition} \\
    \theta_{\text{lock},n} & \hspace*{0mm} \mbox{otherwise}
    \end{cases}
\end{equation}

\begin{equation}
    s_{n+1}= \begin{cases}
        0 & \hspace*{0mm} \mbox{if $n=0$}\\
        1 & \hspace*{0mm} \mbox{if $s_{n}=1$} \label{eq: progress transition} \\
        s_n + \Delta s & \hspace*{0mm}  \mbox{otherwise}
    \end{cases}
\end{equation}
with $\Delta s = 0.01$.

The locking status can change from 0 to 1 depending on action $u_n$.
The locking mechanism automatically unlocks if the deflection angle decreases below zero, as described by \eqref{eq: lock transition}.
The deflection angle is calculated using equation \eqref{eq: deflection angle}.

Evaluation of equation \eqref{eq: deflection angle} requires storing the joint angle at which the support mechanism is locked as long as $l_n=1$, as given by \eqref{eq: theta transition}.

The progress value increases from one time step to another, independently of a controller action, as given by \eqref{eq: progress transition}.

\textbf{Instantaneous reward model:}
The instantaneous reward model captures the impact of the exoskeleton on a user at each instance $s_n$. 
The instantaneous reward is based on the \textit{residual} human load defined as the difference between human joint torques $\tau_{\text{hum}}(s_n)$ and exoskeleton joint torques $\tau_{\text{exo}}(\alpha_n)$.
If the support mechanism is unlocked, i.e. $l_n=0$, it is assumed that the exoskeleton does not increase or reduce human joint torques.
Therefore, the residual load is calculated as:
% Here, a model is chosen based on human joint angle $\theta_{\text{hum}}(s)$ and joint torque $\tau_{\text{hum}}(s)$ values and exoskeleton joint torque values $\tau_{\text{exo}}(\theta_{\text{hum}}(s_k),\theta_{l,k})$.
\begin{equation}
    \tau_{\text{res},n} =
        \begin{cases}
            \tau_{\text{hum}}(s_n)& \text{if $l_n=0$} \\
            \tau_{\text{hum}}(s_n) - \tau_{\text{exo}}(\alpha_n) & \text{if $l_n=1$} \label{eq: residual load}
        \end{cases}
\end{equation}
$\tau_{\text{exo}}(\alpha_n)$ is calculated using the MACCEPA 2.0 actuator model described in section \ref{sec: exoskeleton hardware} and visualized in figure \ref{fig: torqueDeflection}.

The instantaneous reward is calculated as
\begin{equation}
    \label{eq: instantaneous reward}
    \rho_n = -\tau_{\text{res},n}^2
\end{equation}
To conform to reinforcement learning conventions, maximizing the reward should lead to the optimal behavior. 
Therefore, a minus sign is added to equation \eqref{eq: instantaneous reward}. 
Similarly as in equation \eqref{eq: total reward lock}, the residual load is squared to emphasize reductions of peak joint torques.

\textbf{Total reward:}
Value iteration is now used to calculate so-called $Q(x_n,u_n)$-functions which are well-known from reinforcement learning. 
These $Q(x_n,u_n)$-functions represent the total reward obtained from taking an action $u_n$ in state $x_n$ under the assumption that the controller will only take reward-maximizing actions in the future.
The value iteration algorithm iteratively updates these $Q(x_n, u_n)$-functions according to the Bellman equation \cite{Sutton2018}:
% Value iteration is used to calculate the utilities $U(s_{k,j}, a_j|\text{model}_j)$ from equation \eqref{eq: expected utility}.
% For every state $x_k$ and every action $u_k$, a so-called $Q(x_k, u_k)$-function will be initialized to zero.
% At a given state $x_k$ and for a given action $u_k$, $Q(x_k, u_k)$ consists of the current reward $\rho_k$ as well as the maximum possible reward to be obtained in the next state $x_{k+1}$.
% This $Q(x_k,s_k)$-function is the utility function for one side of the exoskeleton.
% The utility functions from equation \eqref{eq: expected utility} are obtained by summing the utility functions for both sides of the exoskeleton.

% The value iteration algorithm will iterate over all states and actions and update this $Q(x_k, u_k)$-function according to the Bellman equation \cite{Sutton2018}:
\begin{equation}
    \label{eq: reinforcement update}
    Q(x_n, u_n) = \rho_n + \epsilon \text{max}_{u_{n+1}}(Q(x_{n+1}, u_{n+1}))
\end{equation}
The max operator in equation \eqref{eq: reinforcement update} imposes that the controller should always take actions that maximize future rewards.
This future reward can be discounted with a value $\epsilon$, but here $\epsilon=1$.
The value iteration algorithm is repeated for $N$ iterations such that the $Q$-function for states with $s_n=0$ can incorporate information about the entire future motion (until $s_N=1$).

The resulting $Q(x_n,u_n)$-functions are a function of the entire state vector.
However, the controller can only influence the state of the exoskeleton if the support mechanism is unlocked i.e. $l_n=0$ (because unlocking happens automatically).
Moreover, whenever the exoskeleton is unlocked (i.e. $l_n=0$), then $\theta_{\text{lock},n}=0$ as well.
Consequently, only $Q(l_n=0,\theta_{\text{lock},n}=0, s_n, u_n)$ will be queried during execution to evaluate whether to lock the support mechanism.
This relevant part of the $Q$-function only depends on the progress and action and is referred to as the total reward $q(s_n,u_n)$. 

For each motion$_j$ and each leg, $q(s_n=s_{\text{curr}},u_n=1)$ is identical to $q(s_{\text{curr}}, \text{lock})$  from equation \eqref{eq: total reward lock}.
For both legs in squat motions and the supporting leg in asymmetric stoop motions  $q(s_n=s_{\text{curr}},u_n=0)$ is identical to $q_{\text{supp.}}(s_{\text{curr}}, \overline{\text{lock}})$ from equation \eqref{eq: total reward unlock supp}.
Finally, for the non-supporting leg in asymmetric stoop motions $q(s_n=s_{\text{curr}},u_n=0)$ is identical to $q_{\text{non-supp.}}(s_{\text{curr}}, \overline{\text{lock}})$ from equation \eqref{eq: total reward unlock non supp}.  

From these total reward functions, the utility functions can be calculated as discussed in section \ref{sec: calculating utilities}.